\documentclass{article} 
\usepackage{iclr2027_conference,times}

\usepackage{amsmath,amsfonts,bm}

\def\eqref#1{equation~\ref{#1}}

\def\1{\bm{1}}

\DeclareMathAlphabet{\mathsfit}{\encodingdefault}{\sfdefault}{m}{sl}
\SetMathAlphabet{\mathsfit}{bold}{\encodingdefault}{\sfdefault}{bx}{n}

\usepackage{hyperref}
\usepackage{url}

\usepackage{amssymb}
\usepackage{algorithm}
\usepackage{algorithmic}
\usepackage[table]{xcolor}
\usepackage{multirow}
\usepackage{graphicx}
\usepackage{wrapfig}
\usepackage{booktabs}
\usepackage{hyperref}
\usepackage{url}

\title{Calibrating Attention-Based Token Pruning for Efficient Multi-Image Synthesis in UMMs}

\author{
\textbf{Zimeng Wu}\textsuperscript{1,2,\S}\quad
\textbf{Qijian Tian}\textsuperscript{3}\quad
\textbf{Penghui Du}\textsuperscript{2}\quad
\textbf{Changqian Yu}\textsuperscript{2,$\ddagger$}\quad
\textbf{Jiaxin Chen}\textsuperscript{1,$\dagger$}\quad
\textbf{Yunhong Wang}\textsuperscript{1,$\dagger$}\\[2pt]
\textsuperscript{1}Beihang University\quad
\textsuperscript{2}KlingAI Research\quad
\textsuperscript{3}Shanghai Jiao Tong University\quad
}

\iclrfinalcopy 
\begin{document}

\vspace*{-20pt}
\maketitle

\begingroup
\renewcommand{\thefootnote}{\fnsymbol{footnote}}
\footnotetext[0]{%
\textsuperscript{$\S$}Work done during internship at KlingAI Research.\quad
\textsuperscript{$\ddagger$}Project leader.\quad
\textsuperscript{$\dagger$}Corresponding authors.
}
\endgroup

\vspace{-10pt}

\begin{abstract}
Unified Multimodal Models (UMMs) enable powerful multi-image synthesis but incur substantial computation from long visual contexts. While token pruning offers a direct route to accelerating inference, existing methods are primarily adapted from understanding tasks, suffering from insufficient reference image exploration, confused conditional guidance, and pooling-induced ambiguity that degrades generation quality. To address these issues, we propose CATP, the first training-free context-compression method tailored to UMM generation, which \textbf{C}alibrates the discrepancies of general \textbf{A}ttention-based \textbf{T}oken \textbf{P}runing for better quality-efficiency trade-off. Concretely, building on independent layer-wise pruning during denoising, we introduce Guidance-Consistent Pruning (GCP) to synchronize pruning decisions across conditioning branches, and a Taylor-Calibrated Attention Metric (TCAM) that derives a theoretically grounded second-order correction for pooling-induced bias. In addition, a Fast Scoring Operator (FSO) fuses the calibration into tiled online Softmax to reduce pruning cost. Extensive experiments across representative UMMs and benchmarks demonstrate that CATP consistently outperforms state-of-the-art token-reduction methods, achieving up to 2.09$\times$ attention-core speedup. Code will be released upon paper acceptance.
\end{abstract}

\vspace{-10pt}

\definecolor{mygreen}{RGB}{92,171,87}

\newcommand{\frameworkfigure}{%
\begin{figure}[t]
\centering
\includegraphics[width=\linewidth]{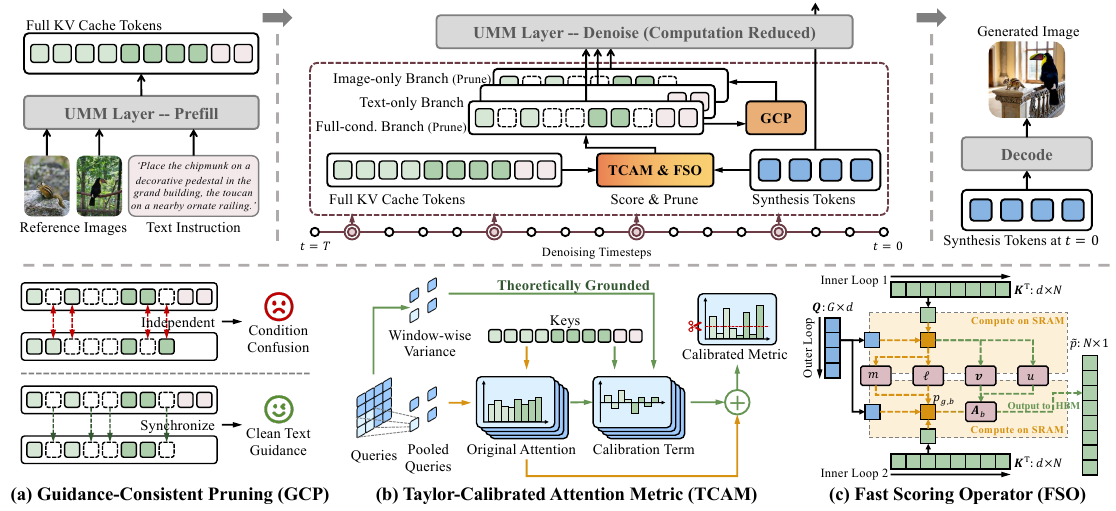}
\vspace{-20pt}
\caption{Framework overview. (Top) UMM inference pipeline with CATP. It accelerates denoising by pruning visual KV cache tokens across layers at predefined timesteps. (Bottom) Illustration of the main modules. (a) GCP synchronizes pruning decisions from the full-condition to image-only branch, avoiding condition confusion from independent pruning. (b) TCAM estimates token importance using pooled-query attention and calibrates it with a theoretically grounded second-order term. (c) FSO efficiently realizes calibrated scoring via a two-pass tiled online-Softmax kernel.}
\vspace{-10pt}
\label{fig:framework}
\end{figure}
}

\newcommand{\generationfigure}{%
\begin{figure}[t]
\centering
\includegraphics[width=\linewidth]{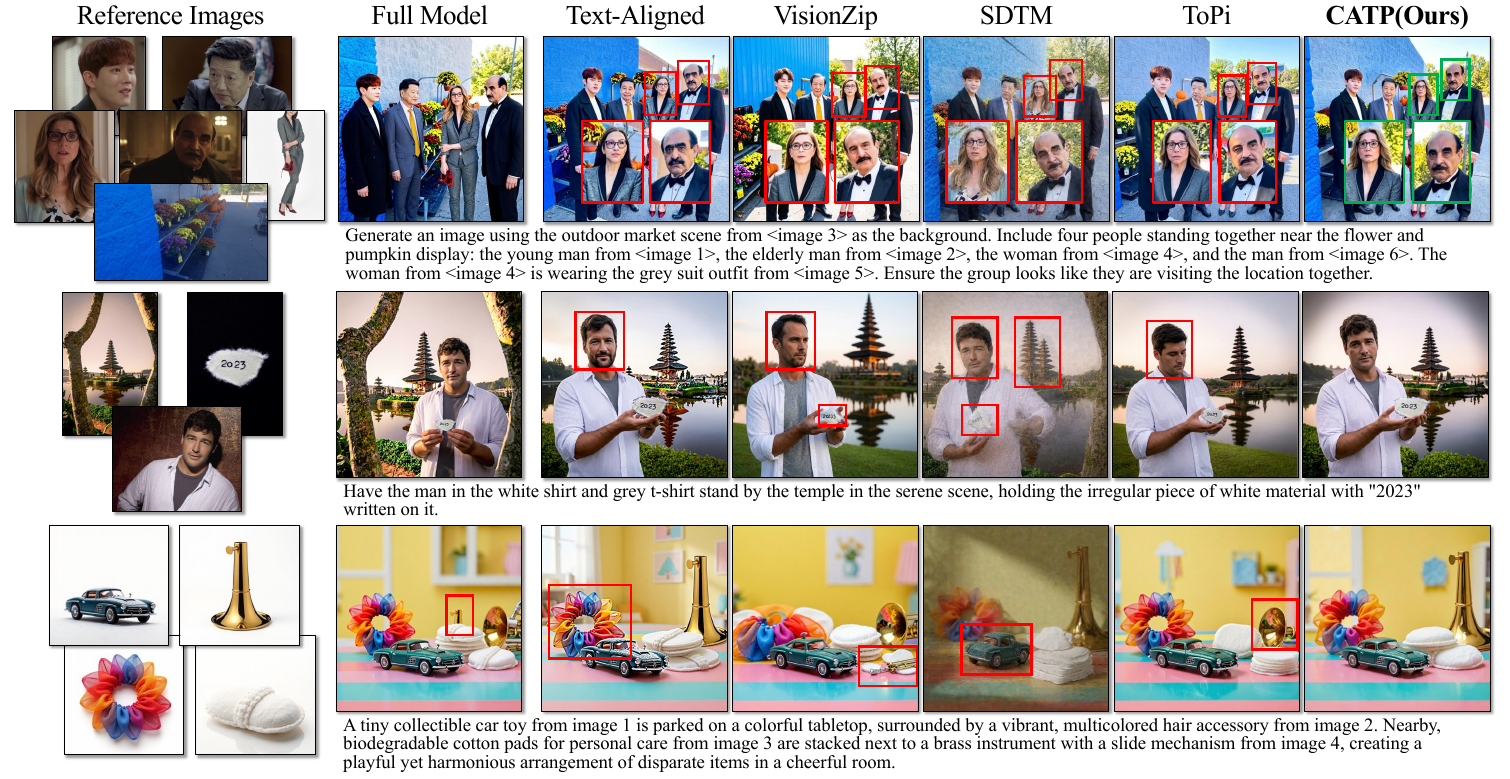}
\vspace{-20pt}
\caption{Qualitative comparison of various methods on Bagel at a 25\% token retention ratio. The failure/well-preserved regions are marked with \textcolor{red}{red}/\textcolor{mygreen}{green} boxes.}
\vspace{-15pt}
\label{fig:generation}
\end{figure}
}

\newcommand{\comparetimefigure}{%
\begin{wrapfigure}[12]{r}{0.5\columnwidth}
\centering
\includegraphics[width=\linewidth]{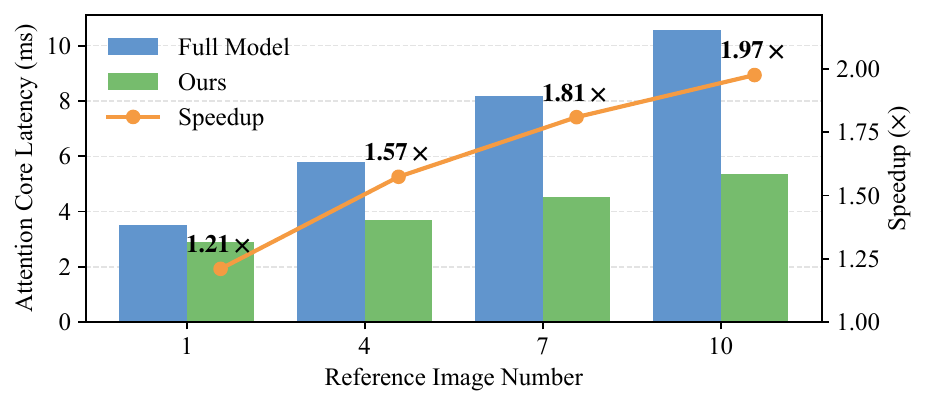}
\vspace{-20pt}
\caption{Attention core latency of the full model and CATP, with relative speedup of CATP, across varying reference image numbers.}
\label{fig:compare_time}
\end{wrapfigure}
}

\newcommand{\motivationfigure}{%
\begin{figure}[t]
\centering
\includegraphics[width=\linewidth]{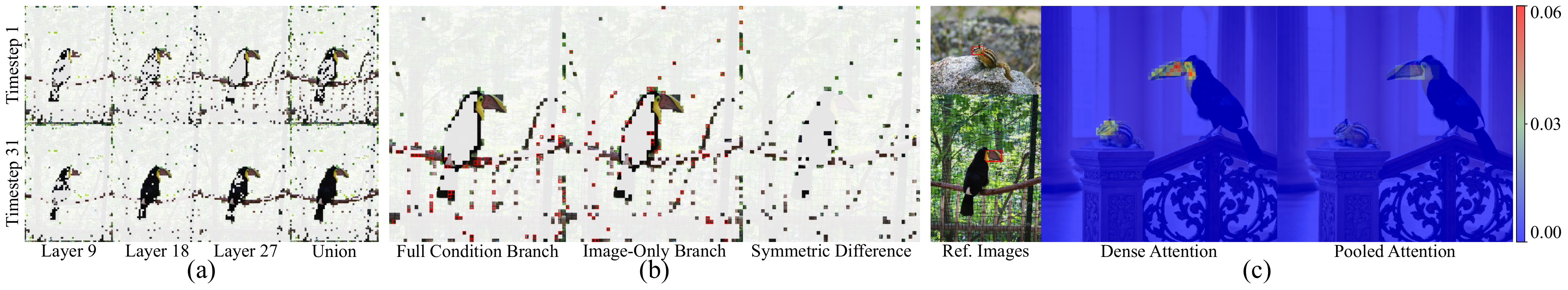}
\vspace{-20pt}
\caption{Empirical observations on Bagel using the general attention metric. (a) Top-10\% tokens at timesteps 1/31 across layers 9/18/27, with their union in the last column. (b) Top-10\% tokens from the (Left) full-condition and (Middle) image-only branches, with (Right) their symmetric difference; red boxes indicate branch-specific tokens. (c) (Left) Reference images with manually selected synthesis-critical regions highlighted in red boxes. (Middle/Right) Heatmaps of the maximum attention from generated-image locations to these regions under dense and pooled attention, respectively.}
\vspace{-10pt}
\label{fig:motivation}
\end{figure}
}

\newcommand{\ablationvisualizefigure}{%
\begin{figure}[!ht]
\centering
\includegraphics[width=\linewidth]{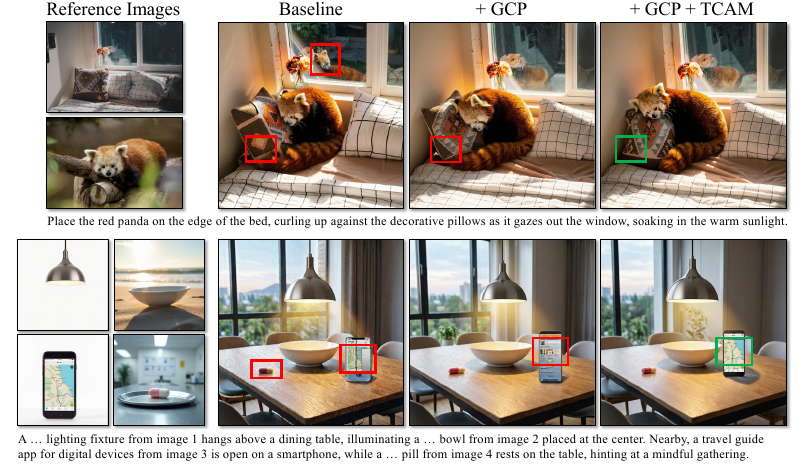}
\vspace{-20pt}
\caption{Qualitative ablation of GCP and TCAM modules on Bagel at a 25\% token retention ratio. The failure/well-preserved regions are marked with \textcolor{red}{red}/\textcolor{mygreen}{green} boxes.}
\label{fig:ablation_visualize}
\end{figure}
}

\newcommand{\timetaylorfigure}{%
\begin{figure}[t]
\centering
\vspace{-10pt}
\includegraphics[width=\linewidth]{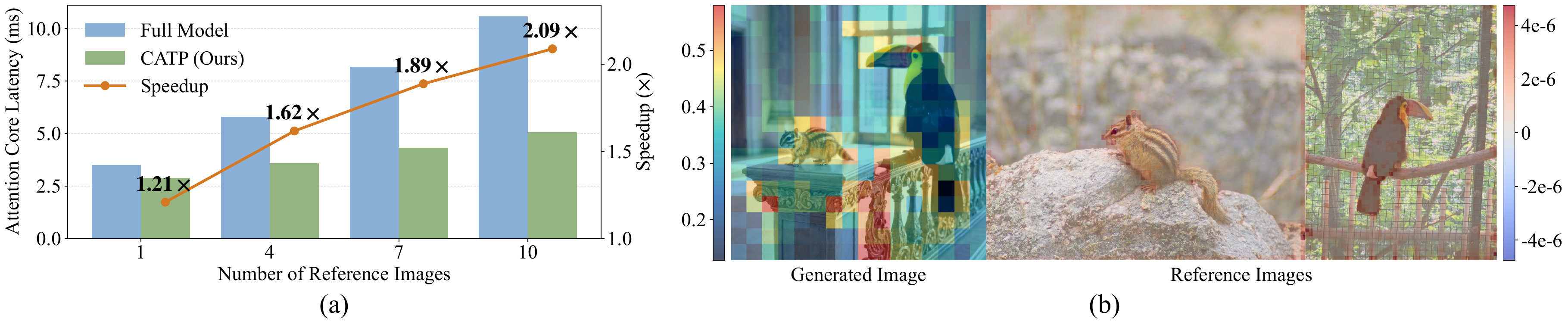}
\vspace{-20pt}
\caption{(a) Attention core latency of the full model and CATP, along with the relative speedup of CATP, across varying numbers of reference images. (b) Visualization of (Left) window-wise query variance and (Right) the resulting aggregated calibration term over reference images in TCAM.}
\label{fig:time_taylor}
\end{figure}
}

\newcommand{\combinationvisualizefigure}{%
\begin{figure}[!ht]
\centering
\includegraphics[width=\linewidth]{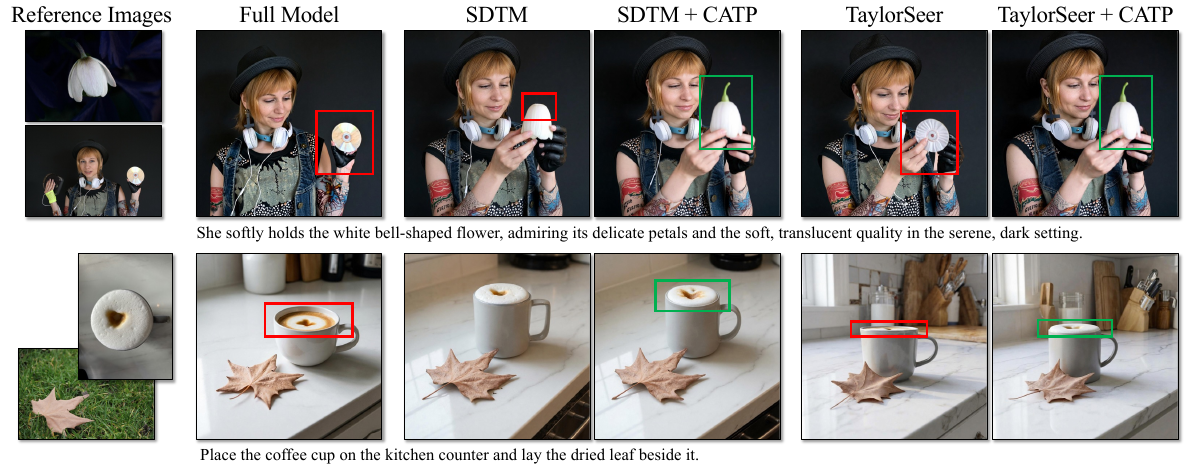}
\vspace{-20pt}
\caption{Qualitative generation results by SDTM and TaylorSeer, individually and combined with CATP, on Bagel. The failure/well-preserved regions are marked with \textcolor{red}{red}/\textcolor{mygreen}{green} boxes.}
\label{fig:combination_visualize}
\end{figure}
}

\section{Introduction}
\label{sec:introduction}
Unified Multimodal Models (UMMs), which integrate understanding and generation within a shared backbone~\citep{bagel,sensenova}, have emerged as a prevalent paradigm for enhancing both capabilities~\citep{ummsurvey}. Leveraging their strength in handling complex inputs, multi-image synthesis, which generates a target image from multiple references~\citep{macro,micobench}, has recently attracted growing interest in broad applications including fashion design \citep{idmvton}, personalized content production \citep{msdiffusion}, etc. 
However, this typically requires concatenating visual tokens from multiple high-resolution images, yielding long sequences that incur substantial attention computation during denoising~\citep{macro,emma}.

In general, such long-context designs are originally intended to support diverse capabilities of foundation models~\citep{bagel}, whereas a specific task may exploit only a subset of the available information~\citep{sparsevlm}. Therefore, reducing the effective sequence length is a direct route to improving inference efficiency~\citep{tome,fastv}. Among strategies, attention-based token pruning offers an effective and general baseline, leveraging native attention score to discard less informative tokens~\citep{tokenprunesurvey,comp}. Nevertheless, existing schemes are primarily developed for understanding-oriented models~\citep{fastv,spts}, while the few direct adaptations to UMMs overlook generation-specific characteristics~\citep{flashunified,macro}, potentially discarding visual regions critical to faithful synthesis.

Specifically, we identify three key limitations: 1) \textbf{Static pruning across timesteps and layers.} Understanding tasks usually produce short responses~\citep{llava}, with most information aggregated into a few tokens at early layers~\citep{spts}, making one-shot pruning throughout inference feasible~\citep{fastv}. In contrast, visual generation attends to complementary regions across denoising timesteps and layers, as in Fig.~\ref{fig:motivation}(a). Especially under aggressive token budgets, a fixed pruning fails to cover the full reference content, yielding insufficient exploration of visual cues. 2) \textbf{Inconsistent compression across conditioning branches.} Classifier-Free Guidance (CFG)~\citep{cfg}, natively instantiated in UMMs~\citep{bagel}, performs multiple forward passes with different input sequences, naturally producing branch-specific attention scores and motivating independent pruning. However, as shown in Fig.~\ref{fig:motivation}(b), this yields mismatched token selections whose symmetric difference carries useless information yet is amplified in the CFG guidance residual, perturbing guided velocity prediction and compromising generation fidelity. 3) \textbf{Ambiguous token selection under pooled query representations.} Adapting pruning to varying context exploration requires frequent importance estimation, while exact attention-based metric with the still-lengthy denoising queries incurs substantial overhead, undermining the goal of acceleration. Average pooling on queries is commonly adopted to reduce this cost~\citep{spargeattention}. However, as observed in Fig.~\ref{fig:motivation}(c), it smooths fine-grained attention responses to critical regions, obscuring their distinction from less relevant content and thus yielding less reliable context retention.

To address these issues, we propose CATP, a novel training-free framework for efficient multi-image synthesis in UMMs. Starting from the general attention-based token pruning, CATP identifies generation-specific discrepancies and calibrates them at both conditioning-branch and importance-estimation levels, thereby improving generation fidelity. Specifically, CATP prunes reference-image tokens in the KV cache across all layers at predefined timesteps, adapting compression to varying reference explorations. Building on this, we introduce Guidance-Consistent Pruning (GCP), which formulates the cross-branch mismatch of independent compression and synchronizes pruning decisions with the full-condition branch to preserve clean guidance. A second-order Taylor expansion then characterizes the bias of attention metric with pooled-query, leading to a Taylor-Calibrated Attention Metric (TCAM) for more reliable identification of critical reference tokens. Additionally, a Fast Scoring Operator (FSO) is developed with a custom Triton kernel. It integrates the calibration into tiled online Softmax computation, substantially reducing the scoring overhead.

In summary, our main contributions are three-fold:

1) We analyze discrepancies in general attention-based token pruning for UMM multi-image synthesis and propose a training-free acceleration framework, dubbed CATP. To the best of our knowledge, CATP makes the first investigation of token pruning specifically tailored to UMM generation.

2) We develop Guidance-Consistent Pruning (GCP), Taylor-Calibrated Attention Metric (TCAM) and Fast Scoring Operator (FSO), which address the rigidity, inconsistency, and ambiguity of general schemes, jointly enabling a flexible, accurate and efficient online context compression.

3) We conduct extensive experiments on representative UMMs and benchmarks, showing that CATP delivers superior generation quality, with up to 2.09$\times$ attention-core acceleration.

\motivationfigure

\section{Related Work}
\label{sec:related_work}
\subsection{Unified Multimodal Models and Multi-Image Synthesis}
Unified Multimodal Models (UMMs) enable understanding and generation within a single architecture~\citep{ummsurvey}. Recent prevalent paradigm adopts a Large Language Model (LLM) as the backbone \citep{ummsurvey}, where text generation remains autoregressive process, while images are synthesized via iterative flow matching in latent~\citep{bagel} or pixel space~\citep{sensenova}. For image generation, conditioning inputs are encoded and concatenated into a token sequence, cached as fixed KV states, and repeatedly attended to by generated visual tokens throughout denoising~\citep{bagel,hunyuanimage3}.
Multi-image synthesis leverages UMMs' strengths in complex instruction understanding and high-quality generation~\citep{micobench,multibanana}. However, multiple reference images substantially lengthen the input sequence, underscoring the practical necessity of efficient inference, especially in attention computation~\citep{macro}.

\subsection{Efficient Inference via Token Reduction}
\textbf{Token Pruning.} Token pruning has presented clear efficiency gains for long-context tasks~\citep{tokenprunesurvey}, exploiting the task-irrelevant redundancy inherent in input sequences~\citep{spts}. Initially explored in understanding tasks, methods develop various criteria for token selection~\citep{visionzip,divprune}. Among them, attention-based pruning offers a strong baseline across LLMs~\citep{spts}, Multimodal Large Language Models (MLLMs)~\citep{fastv}, and the understanding branch of UMM~\citep{flashunified}, accelerating either prefilling through early token removal~\citep{fastv} or decoding via KV cache pruning~\citep{lookm}. 

\textbf{Token Reduction for Visual Generation.} Efficiency in visual generation has been studied from multiple perspectives, such as timestep reduction~\citep{phaseddmd} and cross-step cache reusing~\citep{taylorseer,seer}, while token reduction is orthogonal to them. Most methods originally target Diffusion Transformers (DiTs)~\citep{dit}, typically reducing the computed sequence by merging and subsequently unmerging denoising tokens~\citep{tomesd,sdtm}. ToPi~\citep{topi} more closely follows the aforementioned token pruning paradigm, but shares pruning decisions across layers, overlooking layer-wise differences in information demand. The few attempts on UMMs largely transfer understanding-oriented techniques~\citep{macro}, neglecting generation-specific characteristics. In this work, we systematically analyze these distinctive properties and develop token pruning specifically tailored to UMM generation.
{
\setlength{\abovedisplayskip}{4pt plus 1pt minus 2pt}
\setlength{\belowdisplayskip}{4pt plus 1pt minus 2pt}
\setlength{\abovedisplayshortskip}{0pt minus 2pt}
\setlength{\belowdisplayshortskip}{4pt plus 1pt minus 3pt}

\section{Methodology}
\subsection{Framework Overview}
\label{seq:framework_overview}
In context of mainstream UMMs, multiple reference images and an instruction are encoded into visual and text token sequences $\bm{I}$ and $\bm{T}$, respectively, and concatenated to form a context $\bm{C}\!=\![\bm{I};\bm{T}]$ of length $N$. Prefilling $\bm{C}$ through the $L$-layer, $\theta$-parameterized backbone $f_{\theta}(\cdot)$ produces layer-wise representations $\{\bm{C}^l\}_{l=1}^{L}$, cached as fixed KV states. Starting from noisy visual tokens $\bm{Z}_T$ of length $M$, generation proceeds for $T$ iterative flow-based denoising steps as Eq.~(\ref{eq:denoising}), where the intermediate visual states $\bm{Z}_t^{l}$ attend to $\bm{C}^l$ within the attention blocks at each layer, and $\bm{Z}_0$ is ultimately decoded into the output image.
\begin{equation}
\bm{Z}_{t-1}=f_{\theta}(\bm{Z}_t;t,\{\bm{C}^l\}_{l=1}^L),~ t=T,\ldots,1.
\label{eq:denoising}
\end{equation}

Token pruning shortens $\bm{C}^l$ at a predefined timestep $t^\mathrm{p}$, reducing attention cost in the current and subsequent denoising steps. Motivated by the dynamic reference exploration discussed in Sec.~\ref{sec:introduction}, we adopt the common independent layer-wise pruning~\citep{lookm} as the baseline framework. Specifically, each token $\bm{C}^l_j$ is assigned an importance score $s_j^l$, and the top-$B$ candidates are retained accordingly as:
\begin{equation}
\hat{\bm{C}}^{t\le t^\mathrm{p},l} = \{\bm{C}_j^l | j \in \operatorname{TopK}(\{s_j^{l}\}_{j=1}^N,B)\}.
\end{equation}

As established in prior work~\citep{topi}, the attention from denoising visual tokens to reference tokens effectively reflects the native information flow during image synthesis, providing a principled criterion for token pruning. Formally, let $\bm{Q}\!=\!\{\bm{q}_m\}_{m=1}^{M}$ denote the queries projected from $\bm{Z}^{l}_{t^\mathrm{p}}$, and $\bm{K}\!=\!\{\bm{k}_j\}_{j=1}^N$ the keys of $\bm{C}^l$, with the superscript $l$ omitted for clarity. Accordingly, the baseline importance score $s_j^l$ is defined as the average attention received by token $j$ over all queries:
\begin{equation}
    s_j^l = \frac{1}{M}\sum\nolimits_{\bm{q}\in \bm{Q}} \operatorname{Attn}(\bm{q})_j,\quad \operatorname{Attn}(\bm{q})=\operatorname{Softmax}\left(\frac{\bm{q}\bm{K}^\top}{\sqrt{d}}\right)\in \mathbb{R}^N,
    \label{eq:raw_attention}
\end{equation}
where $d$ denotes head dimension, and we omit the head-wise averaging from the notation for brevity.

Building upon this general framework, CATP focuses on accelerating the computation-dominant denoising process, as illustrated in Fig.~\ref{fig:framework}. By further investigating the aforementioned limitations of vanilla attention-based pruning, we develop GCP to determine $\hat{\bm{C}}^{t\le t^\mathrm{p},l}$ across condition branches and TCAM with FSO to refine $s_j^l$, thereby enabling more accurate and efficient context compression.

\subsection{Guidance-Consistent Pruning}
\label{seq:gcp}
The native CFG mechanism~\citep{cfg} instantiates Eq.~(\ref{eq:denoising}) into three condition branches, each performing a complete forward pass under full, image-only or text-only conditioning:
\begin{equation}
\bm{V}_t^{\mathrm{full}}=v_{\theta}(\bm{Z}_t,t;\bm{I},\bm{T}),\quad
\bm{V}_t^{\mathrm{img}}=v_{\theta}(\bm{Z}_t,t;\bm{I},\varnothing),\quad
\bm{V}_t^{\mathrm{text}}=v_{\theta}(\bm{Z}_t,t;\varnothing,\bm{T}),
\end{equation}
where $\bm{V}$ denotes the velocity field, and $v_\theta(\cdot)$ is the velocity predictor of $f_\theta(\cdot)$. This yields modality-specific guidance residuals $\bm{G}_t^\mathrm{img}\!=\!\bm{V}_t^\mathrm{full}-\bm{V}_t^\mathrm{text}$ and $\bm{G}_t^\mathrm{text}\!=\!\bm{V}_t^\mathrm{full}-\bm{V}_t^\mathrm{img}$, capturing the incremental contribution of each modality. Accordingly, the guided velocity can be equivalently formulated as $\bm{V}_t^\textrm{cfg}$, controlled by model-inherent weights $\lambda^\mathrm{text}$ and $\lambda^\mathrm{img}$, and then used for denoising update:
\begin{equation}
    \bm{V}_t^\mathrm{cfg} = \bm{V}_t^\mathrm{full} + \lambda^\mathrm{img}\bm{G}_t^\mathrm{img} + \lambda^\mathrm{text}\bm{G}_t^\mathrm{text},\quad \bm{Z}_{t-1} = \bm{Z}_t - \Delta t\bm{V}_t^\mathrm{cfg}.
\end{equation}

\frameworkfigure

Notably, reference image tokens, which are the primary pruning targets, are shared by both the full-condition and image-only branches. However, as shown in Fig.~\ref{fig:motivation}(b), the straightforward strategy of pruning each branch using its native online attention scores yields mismatched token selections. Formally, let $\bm{I}^{\mathrm{full}}$ and $\bm{I}^{\mathrm{img}}$ denote the visual tokens retained in the corresponding branches, with the layer index omitted for simplicity. The resulting text-guidance can be decomposed as:
\begin{equation}
\tilde{\bm{G}}_t^{\text{text}} =
\underbrace{
    v_{\theta}(\bm{Z}_t,t;\bm{I}^{\mathrm{full}},\bm{T})-v_{\theta}(\bm{Z}_t,t;\bm{I}^{\mathrm{full}},\varnothing)
}_{\text{clean text guidance}}
+
\underbrace{
    v_{\theta}(\bm{Z}_t,t;\bm{I}^{\mathrm{full}},\varnothing)-v_{\theta}(\bm{Z}_t,t;\bm{I}^{\mathrm{img}},\varnothing)
}_{\bm{\delta}_t^{\mathrm{conf}}}.
\end{equation}
Here, the first term remains clean and valid text guidance, whereas an unexpected condition confusion arises from the second term $\bm{\delta}_t^{\mathrm{conf}}$, as it introduces a spurious guidance signal from the mismatched visual token sets, further amplified by the non-negligible scale $\lambda^\textrm{text}$.

To address this issue, we adopt a simple yet effective strategy that synchronizes pruning across branches sharing the same visual context. Concretely, token importance is estimated from the full-condition branch, which has access to complete context, and the resulting pruning decision is reused by the image-only branch, yielding:
\begin{equation}
\hat{\bm{I}}^{\mathrm{img},t,l}\leftarrow\hat{\bm{I}}^{\mathrm{full},t,l},~\forall~t,l.
\end{equation}
This eliminates the confusion term as $\bm{\delta}_t^{\mathrm{conf}}\!=\!0$, ensuring the clean textual guidance as intended.

\subsection{Taylor-Calibrated Attention Metric}
\label{sec:tcam}
\textbf{Pooling-Induced Bias.}
Without loss of generality, we focus on a single layer, writing $s_j^l$ in Eq.~(\ref{eq:raw_attention}) as $s_j$ for brevity. Although faithful, the sequence length of $\bm{Q}$ incurs non-negligible computation, undermining the acceleration objective. A common remedy is to spatially downsample $\bm{Q}$ \citep{macro,spargeattention}. 
Concretely, $\bm{Q}$ is partitioned into $G=M/w^2$ non-overlapping $w\times w$ windows $\{\mathcal{W}_g\}_{g=1}^{G}$, each represented by an average-pooled query $\bar{\bm{q}}_g = \frac{1}{w^2}\sum_{\bm{q}\in \mathcal{W}_g}\bm{q}$. This replaces the costly scoring over $\bm{Q}$ in Eq.~(\ref{eq:raw_attention}) with a compact query set, yielding $\bar{s}_j = \frac{1}{G}\sum_{g=1}^{G} \operatorname{Attn}(\bar{\bm{q}}_g)_j$.

However, interchanging averaging with the nonlinear Softmax operation introduces a window-wise numerical bias between $s_j$ and $\bar{s}_j$. Considering a single window $\mathcal{W}_g$, the exact discrepancy is:
\begin{equation}
    \Delta_{g,j} = \frac{1}{w^2} \sum\nolimits_{\bm{q}\in \mathcal{W}_g} \operatorname{Attn}(\bm{q})_j - \operatorname{Attn}(\bar{\bm{q}}_g)_j.
    \label{eq:pool_discrepancy}
\end{equation}

\textbf{Taylor-Based Metric Calibration.}
Given Eq.~(\ref{eq:pool_discrepancy}), our goal is to derive a lightweight estimate of $\Delta_{g,j}$ that compensates $\bar{s}_j$ toward $s_j$, thereby improving the accuracy of token selection. 

First, let $\bm{\epsilon}_{g,i} = \bm{q}_{g,i} - \bar{\bm{q}}_g$ denote the deviation of each query $\bm{q}_{g,i}\in\mathcal{W}_g$ from $\bar{\bm{q}}_g$. Due to the spatial locality of visual features~\citep{swintransformer}, $\bm{\epsilon}_{g,i}$ is typically small, motivating a Taylor approximation of $\operatorname{Attn}(\cdot)_j$ around $\bar{\bm{q}}_g$. Defining $p_{g,j}=\operatorname{Attn}(\bar{\bm{q}}_g)_j$, and the second-order expansion gives:
\begin{equation}
\operatorname{Attn}(\bm{q}_{g,i})_j \approx p_{g,j}+\nabla p_{g,j}^{\top}\bm{\epsilon}_{g,i}+\frac{1}{2}\bm{\epsilon}_{g,i}^{\top}\bm{H}_{g,j} \bm{\epsilon}_{g,i},
\label{eq:taylor_expansion}
\end{equation}
where $\nabla p_{g,j}$ and $\bm{H}_{g,j}$ are the gradient vector and Hessian matrix of $\operatorname{Attn}(\cdot)_j$ at $\bar{\bm{q}}_{g}$, respectively.
Subsequently, averaging Eq.~(\ref{eq:taylor_expansion}) over $\mathcal{W}_g$ yields an approximation of Eq.~(\ref{eq:pool_discrepancy}) as:
\begin{equation}
\Delta_{g,j} \approx \frac{1}{2}\operatorname{Tr}(\bm{H}_{g,j}\bm{\Sigma}_g),\quad \bm{\Sigma}_g = \frac{1}{w^2}\sum\nolimits_{i=1}^{w^2}\bm{\epsilon}_{g,i}\bm{\epsilon}_{g,i}^\top,
\label{eq:trace}
\end{equation}
where $\operatorname{Tr}(\cdot)$ denotes the trace of a matrix, and $\bm{\Sigma}_g$ is the covariance matrix of query deviations. 

Furthermore, define the attention-weighted key mean as $\bar{\bm{k}}_g = \sum_{j=1}^N p_{g,j}\bm{k}_j$ and denote the diagonal entries of $\bm{\Sigma}_g$ by $(\sigma_{g,1}^2,\ldots,\sigma_{g,d}^2)$. As detailed in Sec.~\ref{sec:app_taylor_metric_calibration}, $\bm{H}_{g,j}$ admits a closed-form expression in Eq.~(\ref{eq:hessian}), while the weak cross-channel correlations enable $\bm{\Sigma}_g\approx \sigma_{g}^2\bm{I}$, where $\sigma_{g}^2 = \frac{1}{d}\sum\nolimits_{c=1}^d \sigma_{g,c}^2$. 
\begin{equation}
\bm{H}_{g,j}=\frac{p_{g,j}}{d}[(\bm{k}_j - \bar{\bm{k}}_g)(\bm{k}_j - \bar{\bm{k}}_g)^\top - \sum\nolimits_{n=1}^{N} p_{g,n} (\bm{k}_n - \bar{\bm{k}}_g)(\bm{k}_n - \bar{\bm{k}}_g)^\top].
\label{eq:hessian}
\end{equation}
Substituting these results into Eq.~(\ref{eq:trace}) yields an efficient computable approximation:
\begin{equation}
\Delta_{g,j} \approx \frac{p_{g,j}}{2d}(r_{g,j} - \bar{r}_g), \quad r_{g,j} = \sigma_{g}^2 \|\bm{k}_j - \bar{\bm{k}}_g\|_2^2,\quad \bar{r}_g = \sum\nolimits_{n=1}^N p_{g,n} r_{g,n}.
\end{equation}

Finally, aggregating the window-wise corrections formulates the calibrated importance score $\tilde{s}_j$ as Eq.~(\ref{eq:calibration}), where we introduce the hyperparameter $\lambda$ to control the strength of correction. 
\begin{equation}
    \tilde{s}_j = \frac{1}{G}\sum\nolimits_{g=1}^{G} \tilde{p}_{g,j}, \quad \tilde{p}_{g,j}=p_{g,j}+\lambda\frac{p_{g,j}}{2d}(r_{g,j} - \bar{r}_g).
    \label{eq:calibration}
\end{equation}

\emph{Remark.} This derivation indicates that pooling-induced bias is jointly governed by local query variation and key dispersion within the corresponding attention view. Larger query variance implies greater information loss from pooling, while larger centered key distance makes attention response more sensitive to such variation, both requiring stronger correction. This also echoes prior heuristic use of variance for regional importance~\citep{sdtm,ukmp}, while our derivation further reveals its theoretical role in modulating importance rather than directly estimating it.

\subsection{Fast Scoring Operator}
\label{sec:fso}
While parallelizable across windows and keys, the calibration still involves a sequential dependency $p_{g,j}\rightarrow \bar{\bm{k}}_g \rightarrow r_{g,j} \rightarrow \bar{r}_g \rightarrow \tilde{p}_{g,j}$, where the repeated $p_{g,j}$-weighted aggregations require multiple accesses to large-scale matrices, incurring non-negligible memory traffic. Therefore, we develop a tiled window-wise scoring operator that fuses the correction into the native computation of $p_{g,j}$.

First, as detailed in the Sec.~\ref{sec:app_simplified_formula_derivation}, $\tilde{p}_{g,j}$ can be equivalently reformulated as:
\begin{equation}
\tilde{p}_{g,j}=p_{g,j}
[1+\frac{\lambda\sigma_g^2}{2d}
(\|\bm{k}_j\|_2^2-2\bm{k}_j^\top\bar{\bm{k}}_g+2\|\bar{\bm{k}}_g\|_2^2-\sum\nolimits_{n=1}^{N}p_{g,n}\|\bm{k}_n\|_2^2)].
\label{eq:simplified_calibration}
\end{equation}
This avoids the explicit dependence on $\bar{r}_{g}$, allowing the required statistics to be accumulated alongside online Softmax. Concretely, as summarized in Algo.~\ref{alg:tcam_operator}, we adopt a tiled two-pass procedure inspired by FlashAttention~\citep{flashattention2} and Sec.~\ref{sec:app_simplified_formula_derivation}. Notably, without materializing intermediate logits, two passes are minimally required to explicitly recover the normalized probabilities $p_{g,j}$. 

During the first pass, the native logit maximum $m_b$ and normalizer $\ell_b$ are updated online in Static Random-Access Memory (SRAM), and the attention-weighted key mean $\bm{v}$ and squared-norm statistic $u$ are simultaneously accumulated under the same rescaling scheme, as Lines~\ref{line:1}$\sim$\ref{line:2}. After traversing all key blocks, normalization by $\ell_{T_K}$ yields the exact $\bar{\bm{k}}_g$ and $\sum\nolimits_{n=1}^{N}p_{g,n}\|\bm{k}_n\|_2^2=u$, as Line~\ref{line:3}. 

The second pass reloads each key block and recomputes the exact $p_{g,j}$ utilizing the finalized statistics $m_{T_K}$ and $\ell_{T_K}$, as shown in Line~\ref{line:4}. With all window-level statistics available, the correction then depends only on the current key-specific terms $\|\bm{k}_j\|_2^2$ and $\bm{k}_j^\top\bar{\bm{k}}_g$, enabling $\tilde{p}_{g,j}$ to be computed immediately as Line~\ref{line:5} and accumulated back into High Bandwidth Memory (HBM) score tile. 

Overall, only calibrated scores are materialized, avoiding separate attention and correction matrices.

{
\global\textfloatsep=8pt plus 2pt minus 2pt\relax 
\begin{algorithm}[t]
\small
\newcommand{\nativeop}[1]{\textcolor{brown}{#1}}
\newcommand{\taylorop}[1]{\textcolor{green!50!black}{#1}}
\renewcommand{\algorithmicrequire}{\textbf{Input:}}
\renewcommand{\algorithmicensure}{\textbf{Output:}}
\caption{Inner Loop of the Fast Scoring Operator (FSO) (\nativeop{Native attention}; \taylorop{Taylor correction}).}
\label{alg:tcam_operator}
\begin{algorithmic}[1]
\REQUIRE Pooled query $\bar{\bm{q}}_g$, variance $\sigma_g^2$, key matrix $\bm{K}\!\in\!\mathbb{R}^{N\times d}$ in HBM; $\bm{K}$ is partitioned into $T_K\!=\!\lceil N/B_K\rceil$ blocks $\{\bm{K}_b\}_{b=1}^{T_K}$, each of size $B_K\!\times \!d$; hyperparameter $\lambda$; $\frac{1}{\sqrt{d}}$ is absorbed into $\bm{K}$ and omitted for brevity.
\ENSURE Calibrated importance scores $\tilde{\bm{p}}_g\in\mathbb{R}^{N}$ in HBM.
\STATE Load $\bar{\bm{q}}_g, \sigma_g^2$ from HBM to on-chip SRAM.
\STATE On chip, initialize \nativeop{$(m_0,\ell_0)=(-\infty,0)$} and
\taylorop{$(\bm{v},u)=(\bm{0},0)$} in SRAM.
\STATE \emph{\nativeop{Pass 1: online Softmax}
\taylorop{with key-moment accumulation}.}
\FOR{$b = 1,\ldots, T_K$}
    \STATE Load $\bm{K}_b$ from HBM to on-chip SRAM.
    \STATE On chip, compute \nativeop{$\bm{z}_b\leftarrow\bar{\bm{q}}_g\bm{K}_b^\top$, $m_b\leftarrow\max(m_{b-1}, \max(\bm{z}_b))$}.
    \label{line:1}
    \STATE On chip, compute \nativeop{$\bm{e}_b\leftarrow\exp(\bm{z}_b-m_b)$, $\alpha\leftarrow\exp(m_{b-1}-m_b)$, $\ell_b\leftarrow\alpha\ell_{b-1}+\operatorname{sum}(\bm{e}_b)$}.
    \STATE On chip, compute \taylorop{$\bm{v}\leftarrow\alpha\bm{v}+\bm{e}_b\bm{K}_b$, $u\leftarrow\alpha u+\bm{e}_b\operatorname{RowSqNorm}(\bm{K}_b)^\top$}.
    \label{line:2}
\ENDFOR
\STATE On chip, compute \taylorop{$\bar{\bm{k}}_g\leftarrow\bm{v}/\ell_{T_K}$, $u\leftarrow u/\ell_{T_K}$}.
\label{line:3}
\STATE \emph{\nativeop{Pass 2: probability recomputation}
\taylorop{with Taylor calibration}.}
\FOR{$b = 1,\ldots, T_K$}
    \STATE Reload $\bm{K}_b$ from HBM to on-chip SRAM.
    \STATE On chip, compute \nativeop{$\bm{z}_b\leftarrow\bar{\bm{q}}_g\bm{K}_b^\top$, $\bm{p}_{g,b}\leftarrow\exp(\bm{z}_b-m_{T_K})/\ell_{T_K}$}.
    \label{line:4}
    \STATE On chip, compute \taylorop{$\bm{A}_b\leftarrow\operatorname{RowSqNorm}(\bm{K}_b)\!-\!2\bm{\bar{\bm{k}}_g}\bm{K}_b^\top\!+\!2\|\bm{\bar{\bm{k}}_g}\|_2^2\!-\!u$, $\tilde{\bm{p}}_{g,b}\leftarrow\bm{p}_{g,b}\odot(1+\frac{\lambda}{2}\sigma_g^2\bm{A}_b)$}.
    \label{line:5}
    \STATE Accumulate $\tilde{\bm{p}}_{g,b}$ into the $b$-th tile of $\tilde{\bm{p}}_g$ in HBM.
\ENDFOR
\RETURN $\tilde{\bm{p}}_g$
\end{algorithmic}
\end{algorithm}
\AddToHookNext{shipout/after}{\global\textfloatsep=20pt plus 2pt minus 4pt\relax} 
}

}
\section{Experimental Results and Analysis}
\definecolor{TableSection}{gray}{0.92}
\definecolor{TableHighlight}{RGB}{235,246,250}

\newcommand{\included}{\ensuremath{\checkmark}}
\newcommand{\sectionrow}[1]{%
  \rowcolor{TableSection}
  \multicolumn{11}{c}{\textit{#1}}\\
}
\newcommand{\categorysectionrow}[1]{%
  \rowcolor{TableSection}
  \multicolumn{10}{c}{\textit{#1}}\\
}

\newcommand{\maintable}{%
\begin{table}[t]
\vspace{-10pt}
\caption{Comparison of multi-image synthesis performance scores by various methods for Bagel, evaluated on the MacroBench and MICo-Bench datasets. The best results are highlighted in \textbf{bold}.}
\label{tab:main_table}
\begin{center}
\small
\setlength{\tabcolsep}{5.5pt}
\begin{tabular}{l|ccccc|ccccc}
\toprule
\multirow{2}{*}{Method} & \multicolumn{5}{c|}{MacroBench} & \multicolumn{5}{c}{MICo-Bench} \\
\cmidrule(lr){2-6}\cmidrule(lr){7-11}
& 1--3 & 4--5 & 6--7 & 8--10 & Avg.$\uparrow$ & Object & Human & HOI & De\&Re & Overall$\uparrow$ \\
\midrule
Full Model & 8.89 & 8.51 & 7.97 & 7.20 & 8.14 & 47.52 & 40.68 & 42.47 & 53.69 & 46.66 \\
\midrule
\sectionrow{retain 25\% tokens}
Text-Aligned & 8.21 & 8.22 & 8.07 & 7.44 & 7.98 & 52.11 & 35.77 & 46.95 & 52.92 & 47.64 \\
VisionZip & 8.18 & 8.07 & 8.00 & 6.97 & 7.81 & 49.65 & 22.99 & 38.79 & 56.40 & 43.38 \\
SDTM      & 8.13 & 7.99 & 7.68 & 7.50 & 7.83 & 38.96 & 24.79 & 25.86 & 41.52 & 32.90 \\
ToPi      & 8.61 & 8.39 & 8.12 & 7.49 & 8.15 & 53.48 & \textbf{40.17} & 46.00 & \textbf{63.69} & 51.97 \\
\rowcolor{TableHighlight} \textbf{CATP (Ours)} & \textbf{8.76} & \textbf{8.69} & \textbf{8.51} & \textbf{7.87} & \textbf{8.46} & \textbf{54.81} & 39.87 & \textbf{49.49} & 63.61 & \textbf{53.22} \\
\midrule
\sectionrow{retain 10\% tokens}
Text-Aligned & 7.09 & 7.30 & 7.53 & 6.95 & 7.22 & 49.42 & 23.60 & 39.72 & 55.18 & 43.35 \\
VisionZip & 5.44 & 6.02 & 6.25 & 5.70 & 5.85 & 37.43 & 13.92 & 27.58 & 46.95 & 33.00 \\
SDTM      & 7.21 & 7.08 & 6.81 & 6.47 & 6.89 & 31.45 & 15.18 & 18.82 & 28.81 & 23.41 \\
ToPi      & 7.32 & 7.18 & 7.28 & 6.86 & 7.16 & 46.89 & 29.29 & 38.81 & 56.60 & 44.20 \\
\rowcolor{TableHighlight} \textbf{CATP (Ours)} & \textbf{8.40} & \textbf{8.33} & \textbf{8.31} & \textbf{7.54} & \textbf{8.15} & \textbf{52.01} & \textbf{29.51} & \textbf{42.88} & \textbf{58.04} & \textbf{46.84} \\
\bottomrule
\end{tabular}
\end{center}
\vspace{-20pt}
\end{table}
}

\newcommand{\omnitable}{%
\begin{table}[t]
\caption{Comparison of performance scores by various methods for Bagel/SenseNova-U1 on the OmniContext dataset with 25\% tokens retained. The best results are highlighted in \textbf{bold}.}
\label{tab:omnicontext_results}
\begin{center}
\small
\setlength{\tabcolsep}{6.5pt}
\begin{tabular}{l|cc|ccc|ccc|c}
\toprule
\multirow{2}{*}{Method} & \multicolumn{2}{c|}{Single} & \multicolumn{3}{c|}{Multiple} & \multicolumn{3}{c|}{Scene} & \multirow{2}{*}{Avg.$\uparrow$} \\
\cmidrule(lr){2-3}\cmidrule(lr){4-6}\cmidrule(lr){7-9}
& Char. & Obj. & Char. & Obj. & Char.+Obj. & Char. & Obj. & Char.+Obj. & \\
\midrule
\multicolumn{10}{c}{\textit{Bagel (Latent Space)}} \\
\midrule
Full Model   & 8.50 & 9.28 & 7.97 & 9.22 & 8.60 & 7.49 & 8.21 & 8.15 & 8.43 \\
\midrule
Text-Aligned & 8.16 & 8.77 & \textbf{7.56} & \textbf{9.10} & 8.25 & 7.46 & 8.23 & 7.64 & 8.14 \\
VisionZip    & 7.06 & 7.90 & 6.78 & 8.64 & 7.97 & 7.40 & 8.50 & 7.58 & 7.73 \\
SDTM         & 7.16 & 8.32 & 6.44 & 7.10 & 7.56 & 6.32 & 7.45 & 6.97 & 7.16 \\
ToPi         & 6.98 & 8.61 & 6.82 & 8.89 & 8.40 & 7.08 & 8.42 & 8.02 & 7.90 \\
\rowcolor{TableHighlight} \textbf{CATP (Ours)} & \textbf{8.45} & \textbf{9.10} & 7.55 & 8.98 & \textbf{8.59} & \textbf{7.57} & \textbf{8.72} & \textbf{8.26} & \textbf{8.40} \\
\midrule
\multicolumn{10}{c}{\textit{SenseNova-U1 (Pixel Space)}} \\
\midrule
Full Model   & 7.75 & 7.78 & 7.71 & 8.85 & 8.10 & 7.12 & 7.06 & 7.06 & 7.68 \\
\midrule
Text-Aligned & 7.11 & 6.96 & 7.69 & 8.60 & 7.43 & 6.97 & 7.80 & 7.32 & 7.48 \\
VisionZip    & 7.34 & 7.51 & 7.31 & 8.66 & 7.92 & 7.06 & 7.21 & 6.89 & 7.49 \\
SDTM         & 7.02 & 7.73 & 6.84 & 8.01 & 7.46 & 6.89 & 6.69 & 6.22 & 7.11 \\
ToPi         & 6.09 & 6.68 & 6.58 & 7.40 & 7.18 & 6.28 & \textbf{7.94} & 6.81 & 6.87 \\
\rowcolor{TableHighlight} \textbf{CATP (Ours)} & \textbf{7.42} & \textbf{8.06} & \textbf{7.76} & \textbf{8.76} & \textbf{8.40} & \textbf{7.52} & 7.69 & \textbf{7.63} & \textbf{7.91} \\
\bottomrule
\end{tabular}
\end{center}
\vspace{-20pt}
\end{table}
}

\newcommand{\moduleablationtable}{%
\begin{table}[t]
\vspace{-10pt}
\caption{Ablation of GCP and TCAM modules for Bagel on MacroBench at two retention ratios.}
\label{tab:module_ablation}
\begin{center}
\small
\setlength{\tabcolsep}{7pt}
\begin{tabular}{cc|ccccc|ccccc}
\toprule
\multirow{2}{*}{GCP} & \multirow{2}{*}{TCAM} & \multicolumn{5}{c|}{Retain 25\% Tokens} & \multicolumn{5}{c}{Retain 10\% Tokens} \\
\cmidrule(lr){3-7}\cmidrule(lr){8-12}
& & 1--3 & 4--5 & 6--7 & 8--10 & Avg.$\uparrow$ & 1--3 & 4--5 & 6--7 & 8--10 & Avg.$\uparrow$ \\
\midrule
 &         & 8.32 & 8.19 & 8.14 & 7.62 & 8.07 & 6.65 & 6.87 & 7.25 & 6.85 & 6.91 \\
\included &  & 8.75 & 8.65 & 8.36 & 7.45 & 8.30 & 8.23 & 8.28 & \textbf{8.39} & 7.38 & 8.07 \\
\included & \included & \textbf{8.76} & \textbf{8.69} & \textbf{8.51} & \textbf{7.87} & \textbf{8.46} & \textbf{8.40} & \textbf{8.33} & 8.31 & \textbf{7.54} & \textbf{8.15} \\
\bottomrule
\end{tabular}
\end{center}
\vspace{-20pt}
\end{table}
}

\newcommand{\operatorefficiencytable}{%
\begin{wraptable}[10]{r}{0.4\textwidth}
\vspace{-20pt}
 \setlength{\belowcaptionskip}{4pt}
\caption{Effect of TCAM and FSO modules on scoring latency (ms) and MacroBench average score (Avg.).}
\label{tab:operator_efficiency}
\centering
\small
\setlength{\tabcolsep}{3pt}
\begin{tabular}{ccc|r|c}
\toprule
Pooling & TCAM & FSO & Latency$\downarrow$ & Avg.$\uparrow$ \\
\midrule
 & & & 102.97$~~$ & 8.48 \\
 \included & & & 7.24$~~$ & 8.30 \\
\rowcolor[gray]{0.92} \included & \included & & 30.12$~~$ & 8.46 \\
\rowcolor[gray]{0.86} \included & \included & \included & 5.39$~~$ & 8.46 \\
\bottomrule
\end{tabular}
\end{wraptable}
}

\newcommand{\omniappendixtable}{%
\begin{table}[!ht]
\vspace{-10pt}
\caption{Comparison of performance scores by various methods for Bagel/SenseNova-U1 on the OmniContext dataset with 10\% tokens retained. The best results are highlighted in \textbf{bold}.}
\label{tab:omnicontext_results_01}
\begin{center}
\small
\setlength{\tabcolsep}{6.5pt}
\begin{tabular}{l|cc|ccc|ccc|c}
\toprule
\multirow{2}{*}{Method} & \multicolumn{2}{c|}{Single} & \multicolumn{3}{c|}{Multiple} & \multicolumn{3}{c|}{Scene} & \multirow{2}{*}{Avg.$\uparrow$} \\
\cmidrule(lr){2-3}\cmidrule(lr){4-6}\cmidrule(lr){7-9}
& Char. & Obj. & Char. & Obj. & Char.+Obj. & Char. & Obj. & Char.+Obj. & \\
\midrule
\multicolumn{10}{c}{\textit{Bagel (Latent Space)}} \\
\midrule
Full Model   & 8.50 & 9.28 & 7.97 & 9.22 & 8.60 & 7.49 & 8.21 & 8.15 & 8.43 \\
\midrule
Text-Aligned & \textbf{7.06} & 7.36 & 5.17 & \textbf{8.33} & 7.33 & 6.13 & 7.95 & 7.26 & 7.07 \\
VisionZip    & 4.38 & 4.62 & 3.67 & 7.50 & 4.01 & 4.27 & 7.11 & 6.06 & 5.20 \\
SDTM         & 6.57 & 6.23 & 5.89 & 5.77 & 6.68 & 5.31 & 6.31 & 5.75 & 6.06 \\
ToPi         & 5.33 & 6.65 & 4.16 & 7.92 & 6.90 & 5.15 & \textbf{8.20} & 6.89 & 6.38 \\
\rowcolor{TableHighlight} \textbf{CATP (Ours)} & 6.74 & \textbf{8.19} & \textbf{6.78} & 7.94 & \textbf{8.25} & \textbf{6.89} & 8.14 & \textbf{7.61} & \textbf{7.57} \\
\midrule
\multicolumn{10}{c}{\textit{SenseNova-U1 (Pixel Space)}} \\
\midrule
Full Model   & 7.75 & 7.78 & 7.71 & 8.85 & 8.10 & 7.12 & 7.06 & 7.06 & 7.68 \\
\midrule
Text-Aligned & 5.72 & 4.78 & 6.21 & 7.88 & 6.56 & 6.40 & \textbf{7.94} & 6.58 & 6.51 \\
VisionZip    & 6.35 & 6.19 & 5.54 & 7.90 & 6.51 & 5.69 & 7.39 & 6.22 & 6.47 \\
SDTM         & 6.36 & 7.27 & 6.46 & 6.90 & 6.63 & 6.01 & 5.76 & 5.41 & 6.35 \\
ToPi         & 4.70 & 5.64 & 4.45 & 6.96 & 5.38 & 5.13 & 7.54 & 5.87 & 5.71 \\
\rowcolor{TableHighlight} \textbf{CATP (Ours)} & \textbf{7.31} & \textbf{7.45} & \textbf{7.84} & \textbf{8.55} & \textbf{8.18} & \textbf{7.04} & 7.86 & \textbf{7.42} & \textbf{7.71} \\
\bottomrule
\end{tabular}
\end{center}
\vspace{-15pt}
\end{table}
}

\newcommand{\lambdaablationtable}{%
\begin{table}[!ht]
\vspace{-10pt}
\caption{Effect of the correction strength $\lambda$ on MacroBench score for Bagel at a 25\% token retention ratio. The default setting is \underline{underlined}.}
\label{tab:lambda_ablation}
\begin{center}
\small
\setlength{\tabcolsep}{7pt}
\begin{tabular}{c|ccccc}
\toprule
$\lambda$ & 1--3 & 4--5 & 6--7 & 8--10 & Avg.$\uparrow$ \\
\midrule
0    & 8.75 & 8.65 & 8.36 & 7.45 & 8.30 \\
0.25 & 8.73 & 8.70 & 8.34 & 7.73 & 8.38 \\
\underline{0.50} & 8.76 & 8.69 & 8.51 & 7.87 & \textbf{8.46} \\
0.75 & 8.79 & 8.76 & 8.49 & 7.77 & 8.45 \\
1.00 & 8.81 & 8.67 & 8.37 & 7.74 & 8.40 \\
1.25 & 8.83 & 8.65 & 8.35 & 7.68 & 8.38 \\
\bottomrule
\end{tabular}
\end{center}
\vspace{-10pt}
\end{table}
}

\newcommand{\windowablationtable}{%
\begin{table}[!ht]
\vspace{-10pt}
\caption{Effect of the window size $w$ on MacroBench score and scoring latency (ms) for Bagel at a 25\% token retention ratio. The default setting is \underline{underlined}.}
\label{tab:window_ablation}
\begin{center}
\small
\setlength{\tabcolsep}{7pt}
\begin{tabular}{c|ccccc|r}
\toprule
$w$ & 1--3 & 4--5 & 6--7 & 8--10 & Avg.$\uparrow$ & Latency$\downarrow$ \\
\midrule
2 & 8.75 & 8.73 & 8.49 & 7.98 & 8.49 & 17.87$~~~$ \\
\underline{4} & 8.76 & 8.69 & 8.51 & 7.87 & 8.46 & 5.39$~~~$ \\
8 & 8.68 & 8.57 & 8.27 & 7.68 & 8.30 & 2.31$~~~$ \\
\bottomrule
\end{tabular}
\end{center}
\vspace{-10pt}
\end{table}
}

\newcommand{\otherqueriestable}{%
\begin{table}[!ht]
\caption{Effect of various design choices for the pooling-induced bias on MacroBench for Bagel at a 25\% retention ratio. The best results are highlighted in \textbf{bold} and the default setting is \underline{underlined}.}
\label{tab:alternative_scoring}
\begin{center}
\small
\setlength{\tabcolsep}{7pt}
\begin{tabular}{l|ccccc|r}
\toprule
Method & 1--3 & 4--5 & 6--7 & 8--10 & Avg.$\uparrow$ & Latency$\downarrow$ \\
\midrule
Average Pooling & 8.75 & 8.65 & 8.36 & 7.45 & 8.30 & 7.24$~~~$ \\
Max Pooling & 8.59 & 8.56 & 8.29 & 7.50 & 8.24 & 7.23$~~~$ \\
Mixed Query & \textbf{8.80} & \textbf{8.74} & 8.48 & 7.66 & 8.42 & 31.47$~~~$ \\
\underline{TCAM + FSO} & 8.76 & 8.69 & \textbf{8.51} & \textbf{7.87} & \textbf{8.46} & \textbf{5.39}$~~~$ \\
\bottomrule
\end{tabular}
\end{center}
\vspace{-10pt}
\end{table}
}

\newcommand{\moreevaluatortable}{%
\begin{table}[!ht]
\caption{Comparison of multi-image synthesis performance scores by various methods for Bagel on MacroBench, evaluated by Qwen2.5-VL-72B-Instruct and Gemini-3-Flash. The best results are highlighted in \textbf{bold}.}
\label{tab:more_evaluator}
\begin{center}
\small
\setlength{\tabcolsep}{7.0pt}
\begin{tabular}{l|ccccc|ccccc}
\toprule
\multirow{2}{*}{Method} & \multicolumn{5}{c|}{Qwen2.5-VL-72B-Instruct} & \multicolumn{5}{c}{Gemini-3-Flash} \\
\cmidrule(lr){2-6}\cmidrule(lr){7-11}
 & 1--3 & 4--5 & 6--7 & 8--10 & Avg.$\uparrow$ & 1--3 & 4--5 & 6--7 & 8--10 & Avg.$\uparrow$ \\
\midrule
Full Model & 8.43 & 7.65 & 6.45 & 5.42 & 6.99 & 9.01 & 8.48 & 6.93 & 4.45 & 7.21 \\
\midrule
\sectionrow{retain 25\% tokens}
Text-Aligned & 8.07 & 7.52 & 6.26 & 5.34 & 6.80 & 8.05 & 8.34 & 6.76 & 4.82 & 6.99 \\
VisionZip & 7.81 & 7.36 & 5.88 & 5.02 & 6.52 & 7.43 & 7.71 & 6.36 & 3.72 & 6.31 \\
SDTM      & 7.68 & 6.75 & 5.28 & 3.88 & 5.90 & 7.84 & 7.69 & 6.42 & 4.59 & 6.63 \\
ToPi      & 8.10 & 7.45 & \textbf{6.31} & 5.18 & 6.76 & 8.47 & 8.33 & 6.80 & 4.66 & 7.06 \\
\rowcolor{TableHighlight} \textbf{CATP (Ours)} & \textbf{8.44} & \textbf{7.84} & 6.26 & \textbf{5.62} & \textbf{7.04} & \textbf{8.87} & \textbf{8.70} & \textbf{7.54} & \textbf{5.25} & \textbf{7.59} \\
\midrule
\sectionrow{retain 10\% tokens}
Text-Aligned & 7.34 & 6.46 & 5.75 & 4.63 & 6.05 & 6.13 & 6.45 & 5.41 & 3.89 & 5.47 \\
VisionZip & 5.57 & 5.05 & 4.57 & 4.18 & 4.84 & 4.43 & 4.07 & 3.17 & 2.11 & 3.45 \\
SDTM      & 6.71 & 5.40 & 4.08 & 3.05 & 4.81 & 6.74 & 6.37 & 5.42 & 3.79 & 5.58 \\
ToPi      & 7.13 & 6.39 & 5.05 & 4.64 & 5.80 & 6.94 & 6.34 & 4.71 & 3.20 & 5.30 \\
\rowcolor{TableHighlight} \textbf{CATP (Ours)} & \textbf{8.08} & \textbf{7.55} & \textbf{6.03} & \textbf{5.15} & \textbf{6.70} & \textbf{8.10} & \textbf{8.20} & \textbf{7.01} & \textbf{4.99} & \textbf{7.07} \\
\bottomrule
\end{tabular}
\end{center}
\vspace{-10pt}
\end{table}
}

\newcommand{\combinationtable}{%
\begin{table}[!ht]
\caption{MacroBench scores and overall denoising speedup of SDTM and TaylorSeer, individually and in combination with CATP, on Bagel. The best results are highlighted in \textbf{bold}.}
\label{tab:combination_ablation}
\begin{center}
\small
\setlength{\tabcolsep}{7pt}
\begin{tabular}{l|ccccc|c}
\toprule
Method & 1--3 & 4--5 & 6--7 & 8--10 & Avg.$\uparrow$ & Speedup$\uparrow$ \\
\midrule
Full Model & 8.89 & 8.51 & 7.97 & 7.20 & 8.14 & 1.00$\times$ \\
\midrule
 SDTM & \textbf{8.95} & 8.79 & 8.41 & 7.65 & 8.45 & 1.04$\times$ \\
 \textbf{SDTM + CATP (Ours)} & 8.91 & \textbf{8.98} & \textbf{8.63} & \textbf{8.06} & \textbf{8.64} & \textbf{1.41$\times$} \\
\midrule
 TaylorSeer & 8.76 & 8.48 & 8.21 & 7.70 & 8.29 & 2.39$\times$ \\
 \textbf{TaylorSeer + CATP (Ours)} & \textbf{8.78} & \textbf{8.95} & \textbf{8.65} & \textbf{7.81} & \textbf{8.55} & \textbf{3.06$\times$} \\
\bottomrule
\end{tabular}
\end{center}
\vspace{-10pt}
\end{table}
}

\newcommand{\multiseedtable}{%
\begin{table}[!ht]
\caption{Comparison of MacroBench scores (mean $\pm$ std) by various methods on Bagel at a 25\% token retention ratio. The best results are highlighted in \textbf{bold}.}
\label{tab:multi_seed}
\begin{center}
\small
\setlength{\tabcolsep}{7pt}
\begin{tabular}{l|ccccc}
\toprule
Method & 1--3 & 4--5 & 6--7 & 8--10 & Avg.$\uparrow$ \\
\midrule
Full Model   & 8.810 $\pm$ 0.045 & 8.412 $\pm$ 0.083 & 8.048 $\pm$ 0.078 & 7.256 $\pm$ 0.099 & 8.132 $\pm$ 0.047 \\
\midrule
Text-Aligned & 8.210 $\pm$ 0.041 & 8.164 $\pm$ 0.060 & 8.064 $\pm$ 0.048 & 7.406 $\pm$ 0.064 & 7.960 $\pm$ 0.027 \\
VisionZip    & 8.128 $\pm$ 0.065 & 8.010 $\pm$ 0.091 & 7.966 $\pm$ 0.168 & 7.012 $\pm$ 0.153 & 7.780 $\pm$ 0.065 \\
SDTM         & 8.110 $\pm$ 0.222 & 8.048 $\pm$ 0.120 & 7.876 $\pm$ 0.157 & 7.252 $\pm$ 0.165 & 7.822 $\pm$ 0.066 \\
ToPi         & 8.570 $\pm$ 0.056 & 8.278 $\pm$ 0.073 & 8.240 $\pm$ 0.087 & 7.532 $\pm$ 0.113 & 8.152 $\pm$ 0.032 \\
\rowcolor{TableHighlight} \textbf{CATP (Ours)} & \textbf{8.814 $\pm$ 0.073} & \textbf{8.640 $\pm$ 0.104} & \textbf{8.540 $\pm$ 0.039} & \textbf{7.838 $\pm$ 0.058} & \textbf{8.458 $\pm$ 0.020} \\
\bottomrule
\end{tabular}
\end{center}
\vspace{-10pt}
\end{table}
}

\subsection{Experimental Settings}
\textbf{Models.}
Following recent advances in UMMs, we evaluate methods on two representative models: Bagel~\citep{bagel} and SenseNova-U1~\citep{sensenova}, with generation performed in latent and pixel spaces, respectively. For Bagel, we use the multi-image fine-tuned checkpoint from~\citep{macro}, while SenseNova-U1 adopts its official checkpoint with native multi-image support.

\textbf{Datasets and Evaluation Metrics.}
We conduct experiments on three widely used multi-image synthesis benchmarks, following their official splits and metric aggregation rules. Concretely, we report the geometric mean of Image Consistency and Prompt Following scores on the MacroBench~\citep{macro} Customization subset across four reference-number groups (1--3, 4--5, 6--7, and 8--10), the Weighted-Ref-VIEScore for Object-Centric, Human-Centric, Human-Object Interaction (HOI), and Decomposition-and-Recomposition (De\&Re) tasks on MICo-Bench~\citep{micobench}, and the geometric mean of Prompt Following and Subject Consistency scores for the Single/Multiple/Scene categories of OmniContext~\citep{omnigen2}, each further divided into Character (Char.)/Object (Obj.)/Char.+Obj. subcategories. We also include the average score (Avg.) for MacroBench and OmniContext, and the Overall score for MICo-Bench. For metrics involving MLLM-based evaluation, we utilize the open-source Qwen3-VL-32B-Instruct~\citep{qwen3vl} as the evaluator.

\maintable
\omnitable

\textbf{Implementation Details.}
We implement all methods on top of the TorchUMM library~\citep{torchumm}, with FlashAttention~\citep{flashattention2} the acceleration baseline. Since visual tokens dominate the multimodal context, pruning is conducted exclusively on them, including both ViT and VAE tokens in Bagel and patch-level visual tokens in SenseNova-U1. By following~\citep{topi}, pruning is performed over the full sequence at denoising timesteps $t^\mathrm{p}\in\{49,39,29,19,9\}$. Identical retention ratio is applied to each reference image. In all experiments, we fix $\lambda\!=\!0.5$ and $w\!=\!4$ for TCAM and $B_K\!=\!64$ for FSO. All other inference settings, \emph{e.g.}, sampling and guidance parameters, follow the original model configurations. All experiments are conducted on one high-performance GPU.

\subsection{Main Results}
Since token pruning for UMM generation remains rarely explored, we adapt representative baselines from closely related paradigms underlying UMMs, \emph{i.e.}, MLLMs and DiTs. Specifically, Text-Aligned Pruning is previously applied to UMM-based multi-image synthesis by~\citep{macro}. Both it and VisionZip~\citep{visionzip} are originally developed for MLLM understanding tasks. SDTM~\citep{sdtm} and ToPi~\citep{topi} are DiT-oriented methods. For fair comparison, we restrict all methods to attention acceleration and adjust each to an equivalent computation reduction as CATP under predefined token retention ratios.

\textbf{Quantitative Results.} As shown in Tab.~\ref{tab:main_table}, CATP consistently outperforms compared methods across varying numbers of reference images and diverse generation tasks. Specifically, it surpasses the second-best methods by 0.31 and 1.25 points on MacroBench and MICo-Bench, respectively at a 25\% retention ratio, and further increases the advantages to 0.93 and 2.64 points respectively at 10\% retention ratio, demonstrating more pronounced robustness under aggressive context compression. 

Notably, CATP even outperforms the original full model on MacroBench Avg. at a 25\% retention ratio, with gains primarily arising in multi-image settings ($\ge$ 6 images) and reaching up to 0.67 points. This stems from the effective redundancy removal. As discussed in prior work, the intrinsic normalization property of attention can dilute individual weights in excessively long sequences~\citep{spts}, making critical reference regions harder to distinguish and thereby yielding confused spatial layouts or degraded fine-grained details. Accordingly, pruning redundant context mitigates this effect by reducing unnecessary attention competition~\citep{macro}, while CATP further amplifies this benefit by more accurately retaining generation-critical regions.

\textbf{Generalization Across Denoising Spaces.} Given recent advances in pixel-space UMMs, we further adapt methods to the representative SenseNova-U1 and evaluate on OmniContext. As shown in Tab.~\ref{tab:omnicontext_results}, CATP maintains notable advantages of 0.26 and 0.42 Avg. points on Bagel and SenseNova-U1 at 25\% token retention, respectively, with the margins further widening to 0.5 and 1.2 at 10\% retention in Tab.~\ref{tab:omnicontext_results_01}. These results demonstrate its impressive generalizability across denoising spaces.

\generationfigure

\textbf{Qualitative Results.} As shown in Fig.~\ref{fig:generation}, CATP delivers more favorable overall generation quality, particularly in spatial composition, human identity preservation, and critical background details. This stems from the joint effects of our designs, where GCP improves generation fidelity through coherent guidance, and TCAM better preserves fine-grained identity cues by assigning stronger importance corrections to regions with more details, making them more likely to survive pruning.

\textit{Further Remarks.} Text-Aligned relies on attention signals from text or within reference images, which primarily reflect multimodal understanding rather than direct generation relevance. VisionZip merges similar visual regions, favoring coarse structural preservation but potentially losing identity-critical details. SDTM compresses only denoising tokens, making it better suited to simple generation scenarios but prone to substantial degradation when targeting higher acceleration. While ToPi explicitly targets context compression for generation, it reuses pruning decisions across layers to reduce scoring overhead, overlooking layer-wise differences in context exploration, and the fully computed anchor timesteps further limit the compression scale. In contrast, CATP develops a more accurate and efficient scoring process, enabling layer-wise token evaluation and selection. Under aggressive compression, different layers can exploit complementary critical regions rather than maintaining complete target content throughout, thereby maintaining overall generation quality.

\textbf{Efficiency.}
We further assess the acceleration effect of CATP, with FlashAttention as the baseline. For each image count, we fix 1024-resolution inputs, sample 10 examples and report the average latency of the accelerated attention cores across all denoising calls. As shown in Fig.~\ref{fig:time_taylor}(a), attention computation becomes increasingly dominant with more reference images, translating context compression into larger practical gains, reaching 2.09$\times$ with 10 reference images at 25\% retention.

\subsection{Ablation Study}
\timetaylorfigure

\moduleablationtable
\textbf{On GCP.} As shown in Tab.~\ref{tab:module_ablation}, GCP effectively improves conditional-guidance consistency, yielding average gains of 0.23 and 1.16 points at 25\% and 10\% retention ratios, respectively. Intuitively, more aggressive pruning makes the symmetric difference between conditioning branches increasingly dominated by less important tokens, introducing stronger irrelevant signals into the guidance residual. Therefore, GCP becomes more beneficial at lower retention ratios.

\textbf{On TCAM.} Tab.~\ref{tab:module_ablation} also indicates that TCAM further improves gains under both settings. As visualized in Fig.~\ref{fig:time_taylor}(b), high-variance denoising regions concentrate around object boundaries and identity-critical details, consistent with the widely recognized properties in prior studies~\citep{sdtm}. TCAM exploits this signal via second-order calibration to strengthen corresponding reference regions, thus promoting the retention of identity-critical tokens for finer-grained synthesis.

\operatorefficiencytable

\textbf{On FSO.}
We further assess the effect of FSO in Tab.~\ref{tab:operator_efficiency}, by measuring scoring latency with ten 1024-resolution reference images. Concretely, dense attention achieves the best generation quality but incurs prohibitive pruning cost, while query pooling substantially reduces latency at the cost of degraded performance. Given TCAM's additional calibration cost, FSO finally mitigates this by algebraically reformulating it into a fused operator, reducing latency to 5.39 ms without sacrificing quality. Notably, FSO is even faster than the pooling baseline, which owes to fine-grained HBM-level optimization enabling BF16 multiplication with FP32 accumulation.

\section{Conclusion}
In this work, we conduct a pioneering investigation of visual token pruning for efficient UMM multi-image synthesis, and develop a novel training-free method, namely CATP. By performing layer-wise pruning on KV cache tokens, CATP enables adaptive utilization of different reference information. Furthermore, we explicitly preserve clean conditional guidance through Guidance-Consistent Pruning (GCP) and improve the discriminability of critical regions with a theoretically derived Taylor-Calibrated Attention Metric (TCAM), together with a Fast Scoring Operator (FSO) that reduces the scoring cost. Experiments on prevalent latent- and pixel-space UMMs across three multi-image benchmarks demonstrate the effectiveness of CATP, consistently outperforming competitive token-reduction approaches across retention ratios while achieving up to 2.09$\times$ attention-core speedup.

\clearpage
\subsection*{AI use statement}
In this work, we used generative AI tools for assist in the writing of proofs, assist with translation. We have not used generative AI tools for help develop theoretical models or conceptual frameworks, formulate mathematical claims, provide critical ingredients for proving mathematical claims, propose or refine hypotheses, design or provide feedback on research methodology or experiments, implement methods, interpret results, and generate synthetic data sets, clean and reformat dataset, support qualitative and thematic data analysis are not applicable to this work. Additionally, we used generative AI tools for sourcing/searching for information, identify relevant literature. 

We have reviewed all AI-assisted work. Specifically, we independently verified all AI-polished mathematical formulations and proof-related content step by step to ensure their correctness and consistency with the underlying mathematical definitions. AI-assisted translations were manually reviewed by 2 authors to preserve the intended meaning. Information obtained through AI-assisted search was cross-checked against reliable sources, and all AI-identified literature was verified by consulting the corresponding original publications before being cited or discussed in this manuscript. 

We take responsibility for the final content of this work, including text, claims or artifacts produced with the aid of generative AI.

\bibliography{iclr2027_conference}
\bibliographystyle{iclr2027_conference}

\clearpage
\appendix
\numberwithin{equation}{section}
\setcounter{figure}{0}
\setcounter{table}{0}
\renewcommand{\thefigure}{\Alph{figure}}
\renewcommand{\thetable}{\Alph{table}}

\section{More Experimental Results and Analysis}
\subsection{Additional Quantitative Results on OmniContext}
To complement the 25\% retention results reported in Tab.~\ref{tab:omnicontext_results}, we further evaluate all methods on OmniContext~\citep{omnigen2} under a more aggressive 10\% token retention ratio. As shown in Tab.~\ref{tab:omnicontext_results_01}, CATP continues to achieve superior performance in most scenarios on both Bagel and SenseNova-U1, further widening its margin over the second-best methods to 0.5 and 1.2 points, respectively. These results provide additional evidence for the robustness of CATP under aggressive acceleration and its generalization across various UMM generation spaces.
\omniappendixtable

\subsection{Further Ablations}
\textbf{On Calibration Strength $\lambda$.} Since TCAM adopts approximations to avoid computing higher-order terms, its second-order estimate captures the correction trend rather than an exact discrepancy. Therefore, we heuristically introduce $\lambda$ in Eq.~(\ref{eq:calibration}) to control the calibration strength, whose effect is ablated in Tab.~\ref{tab:lambda_ablation}. As $\lambda$ decreases, TCAM approaches the vanilla pooling baseline and provides insufficient emphasis on fine-grained reference regions, whereas an excessively large value biases selection toward high-variance regions and may suppress smoother yet relevant content. Given the overall performance peak at $\lambda=0.5$, we empirically fix this value for all experiments.
\lambdaablationtable

\textbf{On Window Size $w$.} We assess the effect of the pooling window size $w$ in TCAM. As reported in Tab.~\ref{tab:window_ablation}, smaller windows incur less loss of fine-grained cues and therefore yield better generation quality, but at the cost of a longer query sequence with higher scoring latency. Larger windows exhibit the opposite trade-off, reducing scoring overhead while sacrificing more local details. Accordingly, we empirically choose $w=4$ as a balanced configuration and fix it across all experiments.
\windowablationtable

\textbf{Qualitative Ablation on Main Modules.} Beyond the quantitative ablations in Tab.~\ref{tab:module_ablation}, we further visualize the effects of GCP and TCAM in Fig.~\ref{fig:ablation_visualize}. With the general pruning scheme, independent token selection across conditioning branches amplifies irrelevant visual cues through CFG guidance, leading to implausible rendering or distorted content. GCP alleviates this issue by enforcing consistent pruning across branches. TCAM further improves generation by emphasizing information-rich reference regions, yielding local patterns and details that better match the references.

\ablationvisualizefigure

\textbf{On Alternatives for Pooling-Induced Bias.} For the pooling-induced bias discussed in Sec.~\ref{sec:tcam}, we further compare our solution (\emph{i.e.} TCAM + FSO) with several alternative designs in Tab.~\ref{tab:alternative_scoring}. intuitively, the inferior performance of the baseline Average Pooling stems from indiscriminate within-window averaging that smooths fine-grained cues. Therefore, a direct alternative is Max Pooling, however, channel-wise maxima may originate from different spatial tokens, disrupting the semantic coherence of the pooled representation and yielding even worse performance. Inspired by our derivation in Eq.~(\ref{eq:calibration}), which suggests strengthening the contribution of high-variance denoising regions, we further consider an intuitive alternative that leaves queries from the highest-variance windows unpooled, denoted as Mixed Query. Its improved generation quality in Tab.~\ref{tab:alternative_scoring} provides additional empirical support for our motivation, but the enlarged query sequence incurs increased scoring latency. In contrast, TCAM together with FSO achieves gains in both generation quality and scoring efficiency, further validating the effectiveness of our designs.
\otherqueriestable

\subsection{Robustness Across MLLM Evaluators}
Noting that MLLM-based evaluation can exhibit evaluator-specific biases and judgment inconsistencies~\citep{mllmasjudge}, we therefore re-evaluate all methods on MacroBench utilizing two other representative MLLMs: the open-source Qwen2.5-VL-72B-Instruct~\citep{qwen25vl} and the closed-source Gemini-3-Flash~\citep{gemini3flash}. As shown in Tab.~\ref{tab:more_evaluator}, CATP still consistently achieves the best performance across both evaluators and distinct token retention ratios, despite noticeable differences in their scoring distributions. This further suggests that CATP more reliably preserves generation quality rather than benefiting from a particular evaluator. 

Additionally, it is noteworthy that, under the more capable evaluator, performance with larger numbers of reference images remains far from the maximum score (\emph{i.e.} 10.0). While context compression provides partial recovery, we consider that further gains in such challenging scenarios should primarily stem from stronger native multi-reference generation capabilities of the foundation model, motivating continued research along this direction.
\moreevaluatortable

\subsection{Statistical Stability}
We further assess the statistical stability of methods on Bagel at a 25\% token retention ratio. Specifically, all methods are evaluated on MacroBench over five random seeds for image generation, with the mean and standard deviation reported. As shown in Tab.~\ref{tab:multi_seed}, CATP consistently maintains superior performance, improving the Avg. over the second-best method by 0.306 points while exhibiting the lowest Avg. standard deviation of 0.02. These results further demonstrate both the effectiveness and stability of our method.
\multiseedtable

\subsection{Compatibility with Orthogonal Acceleration Methods}
CATP accelerates inference from the perspective of context compression, reducing attention computation by removing redundant conditional tokens. As discussed in Sec.~\ref{sec:related_work}, several methods optimize efficiency along orthogonal dimensions, motivating an exploration of their joint effect with CATP. Specifically, we combine CATP at a 25\% token retention ratio with two representative approaches, SDTM~\citep{sdtm} and TaylorSeer~\citep{taylorseer}, both adopting their official default settings. SDTM shortens the denoising sequence, whereas TaylorSeer performs block-wise computation skipping across timesteps, and we therefore report the end-to-end speedup ratio of the entire denoising process relative to the full model, with 10 input reference images. 

As shown in Tab.~\ref{tab:combination_ablation}, both methods alone already improve average generation quality over the full model. Here, SDTM fails to provide a meaningful acceleration, as the fixed context in UMM paradigm renders computation over its targeted denoising sequence non-dominant, leaving the overhead of frequent selection operation insufficiently amortized. Furthermore, when combined with these methods, CATP continues to reduce computational cost, while even improving generation performance by 0.19 and 0.26 points, respectively, primarily due to the elimination of interfering contextual redundancy. We also include representative generation results in Fig.~\ref{fig:combination_visualize}, where both methods produce erroneous outputs, yet their combination with CATP corrects these failures. These results demonstrate the favorable adaptability of CATP to prevalent techniques, and suggest that appropriate compression may itself provide a promising direction for improving downstream generation quality.
\combinationtable
\combinationvisualizefigure

\section{Detailed Derivations}
\label{app:detailed_derivations}
In this section, we present detailed derivations for TCAM and FSO. Sec.~\ref{sec:app_taylor_metric_calibration} first formulates the pooling-induced attention discrepancy, then derives its second-order Taylor expansion with combined Hessian and query variance, yielding the calibrated importance score. Sec.~\ref{sec:app_simplified_formula_derivation} reformulates this calibration into a simplified form, follow by a clarification of the theoretical basis of exact tiled online Softmax and the analogous online accumulation of the required intermediate statistics.

\subsection{Derivation of the Taylor Calibration}
\label{sec:app_taylor_metric_calibration}

\textbf{Pooling-Induced Bias.}
We begin with the exact attention-based metric $s_j$ computed over the complete query sequence $\bm{Q}$, corresponding to Eq.~(\ref{eq:raw_attention}) in the main text:
\begin{equation}
    s_j=\frac{1}{M}\sum\nolimits_{\bm{q}\in \bm{Q}}\operatorname{Attn}(\bm q)_j, \quad \operatorname{Attn}(\bm{q})=\operatorname{Softmax}(\frac{\bm{q}\bm{K}^\top}{\sqrt{d}})\in \mathbb{R}^N.
    \label{eq:app_exact_attention}
\end{equation}

Afterward, partitioning $\bm{Q}$ into $G=M/w^2$ non-overlapping $w\times w$ windows and applying average pooling yields, for the $g$-th window,
\begin{equation}
    \bar{\bm{q}}_g = \frac{1}{w^2} \sum\nolimits_{i=1}^{w^2} \bm{q}_{g,i},
    \label{eq:app_pooled_q}
\end{equation}
which further produces the coarsened window-level score:
\begin{equation}
    p_{g,j} = \operatorname{Attn}(\bar{\bm{q}}_g)_j.
    \label{eq:app_attn_on_pooled_q}
\end{equation}

Accordingly, grouping the queries by window, the original dense attention metric of Eq.~(\ref{eq:app_exact_attention}) can be equivalently written as:
\begin{equation}
    s_j=\frac{1}{G}\sum\nolimits_{g=1}^{G} \hat{p}_{g,j},\quad \hat{p}_{g,j} = \frac{1}{w^2}\sum\nolimits_{i=1}^{w^2}\operatorname{Attn}(\bm{q}_{g,i})_j,
    \label{eq:app_window_exact_attention}
\end{equation}
whereas the metric with pooled queries becomes:
\begin{equation}
    \bar{s}_j = \frac{1}{G}\sum\nolimits_{g=1}^G \operatorname{Attn}(\bar{\bm{q}}_g)_j = \frac{1}{G}\sum\nolimits_{g=1}^G p_{g,j}.
    \label{eq:app_pooled_attention}
\end{equation}

Although $\bar{s}_j$ can be computed more efficiently, the difference from the exact $s_j$ goes beyond a simple reordering of window-wise averaging and Softmax. Due to the nonlinearity of Softmax, each window incurs an additional non-negligible bias:
\begin{equation}
    \Delta_{g,j} = \hat{p}_{g,j} - p_{g,j} = \frac{1}{w^2}\sum\nolimits_{i=1}^{w^2}\operatorname{Attn}(\bm{q}_{g,i})_j-p_{g,j},
\end{equation}
which is identical to Eq.~(\ref{eq:pool_discrepancy}) of Sec.~\ref{sec:tcam}, and aggregating these window-level discrepancies gives:
\begin{equation}
    s_j = \bar{s}_j + \frac{1}{G}\sum\nolimits_{g=1}^G\Delta_{g,j}.
    \label{eq:app_calibration_target}
\end{equation}
Therefore, recovering the accurate importance score reduces to efficiently estimating the window-level correction $\Delta_{g,j}$, which motivates the following derivation.

\textbf{Taylor Expansion for Discrepancy Approximation.} For the $g$-th window, denote the deviation of each query $\bm{q}_{g,i}$ from the average-pooled representation $\bar{\bm{q}}_g$ as:
\begin{equation}
    \bm{\epsilon}_{g,i}=\bm{q}_{g,i}-\bar{\bm{q}}_g,
\end{equation}
Given the definition in Eq.~(\ref{eq:app_pooled_q}), these deviations sum to zero within the corresponding window:
\begin{equation}
    \sum\nolimits_{i=1}^{w^2}\bm{\epsilon}_{g,i} = \sum\nolimits_{i=1}^{w^2}\bm{q}_{g,i} - w^2 \bar{\bm{q}}_g =\bm{0}.
    \label{eq:app_mean_zero}
\end{equation}

Due to the spatial locality of visual representations, queries within each window typically lie close to their pooled center. This motivates a local Taylor expansion of $\operatorname{Attn}(\bm{q}_{g,i})_{j}$ around $\bar{\bm{q}}_g$. Using Eq.~(\ref{eq:app_attn_on_pooled_q}), the expansion simplifies to:
\begin{equation}
\begin{split}
    \operatorname{Attn}(\bm{q}_{g,i})_j = \operatorname{Attn}(\bar{\bm{q}}_g + \bm{\epsilon}_{g,i})_j & = p_{g,j} +\nabla p_{g,j}^{\top}\bm{\epsilon}_{g,i} +\frac{1}{2}\bm{\epsilon}_{g,i}^{\top}\bm{H}_{g,j}\bm{\epsilon}_{g,i}+\bm{R}^{(3)} \\
    & \approx p_{g,j} +\nabla p_{g,j}^{\top}\bm{\epsilon}_{g,i} +\frac{1}{2}\bm{\epsilon}_{g,i}^{\top}\bm{H}_{g,j}\bm{\epsilon}_{g,i},
\end{split}
\label{eq:app_taylor_expansion}
\end{equation}
where $\nabla p_{g,j} = \nabla_{\bm{q}} \operatorname{Attn}(\bm{q})_j|_{\bm{q}=\bar{\bm{q}}_g}$ and $\bm{H}_{g,j}=\nabla_{\bm{q}}^2\operatorname{Attn}(\bm{q})_j|_{\bm{q}=\bar{\bm{q}}_g}$ denote the first-order derivative and the second-order Hessian matrix, respectively. Terms beyond the second order, represented by $\bm{R}^{(3)}$, are omitted due to their negligible magnitude.

Furthermore, subtracting $p_{g,j}$ from both sides of Eq.~(\ref{eq:app_taylor_expansion}) and averaging over all queries within the window yields:
\begin{equation}
\begin{split}
    \Delta_{g,j} &= \frac{1}{w^2}\sum\nolimits_{i=1}^{w^2} (\operatorname{Attn}(\bm{q}_{g,i})_j - p_{g,j}) \\
    &\approx \frac{1}{w^2}\sum\nolimits_{i=1}^{w^2} (\nabla p_{g,j}^{\top}\bm{\epsilon}_{g,i}) + \frac{1}{w^2}\sum\nolimits_{i=1}^{w^2}(\frac{1}{2}\bm{\epsilon}_{g,i}^{\top}\bm{H}_{g,j}\bm{\epsilon}_{g,i}) \\
    &= \nabla p_{g,j}^{\top}\frac{1}{w^2}\sum\nolimits_{i=1}^{w^2}\bm{\epsilon}_{g,i} + \frac{1}{2w^2}\sum\nolimits_{i=1}^{w^2}(\bm{\epsilon}_{g,i}^{\top}\bm{H}_{g,j}\bm{\epsilon}_{g,i}).
\end{split}
\label{eq:app_discrepancy_quadratic}
\end{equation}

Given Eq.~(\ref{eq:app_mean_zero}), the first-order term vanishes exactly. Meanwhile, by defining the population covariance of the within-window query deviations as:
\begin{equation}
    \bm{\Sigma}_g=\frac{1}{w^2}\sum\nolimits_{i=1}^{w^2}
    \bm{\epsilon}_{g,i}\bm{\epsilon}_{g,i}^{\top},
    \label{eq:app_query_covariance}
\end{equation}
and applying the identity $\bm{x}^{\top}\bm{A}\bm{x}=\operatorname{Tr}(\bm{A}\bm{x}\bm{x}^{\top})$, where $\mathrm{Tr}(\cdot)$ is the matrix trace, the second-order term can be rewritten as $\frac{1}{2}\operatorname{Tr}(\bm{H}_{g,j}\bm{\Sigma}_g)$. Therefore, Eq.~(\ref{eq:app_discrepancy_quadratic}) yields 
\begin{equation}
    \Delta_{g,j}
    \approx\frac{1}{2}\operatorname{Tr}
    (\bm{H}_{g,j}\bm{\Sigma}_g).
    \label{eq:app_trace_approximation}
\end{equation}

\textbf{Hessian of the Softmax Probability.}
For notational convenience, we denote the attention logit associated with $\bm{k}_j$ as:
\begin{equation}
    z_j(\bm{q}) = \frac{\bm{q}^{\top}\bm{k}_j}{\sqrt{d}},
    \label{eq:app_logits}
\end{equation}
such that the Softmax-normalized attention probability in Eq.~(\ref{eq:app_exact_attention}) becomes:
\begin{equation}
    p_j \equiv \operatorname{Attn}(\bm{q})_j=\frac{\exp(z_j(\bm{q}))}{\sum_{n=1}^{N}\exp(z_n(\bm{q}))}.
\end{equation}
Here for brevity, we temporarily omit the window index $g$ and evaluate all quantities at a generic query $\bm{q}$. The derivative of Softmax \emph{w.r.t} its logits is:
\begin{equation}
    \frac{\partial p_j}{\partial z_n}
    =p_j(\mathbb{I}(j=n)-p_n),
    \label{eq:app_softmax_logit_derivative}
\end{equation}
where $\mathbb{I}(\cdot)$ is the indicator function. Applying the chain rule to Eq.~(\ref{eq:app_softmax_logit_derivative}) with Eq.~(\ref{eq:app_logits}) gives:
\begin{equation}
\begin{split}
    \nabla_{\bm{q}}p_j&=\sum\nolimits_{n=1}^{N}\frac{\partial p_j}{\partial z_n}\nabla_{\bm{q}}z_n =\sum\nolimits_{n=1}^N\frac{\bm{k}_n}{\sqrt{d}}p_j(\mathbb{I}(j=n)-p_n)\\
    &=\frac{p_j}{\sqrt{d}}(\bm{k}_j-\sum\nolimits_{n=1}^{N}p_n\bm{k}_n) \\
    &\equiv \frac{p_j}{\sqrt{d}}\bm{c}_j,
\end{split}
\label{eq:app_attention_gradient}
\end{equation}
where we define the attention-weighted mean key $\bar{\bm{k}}$ and the centered key vectors $\bm{c}_j$ as:
\begin{equation}
    \bar{\bm{k}}=\sum\nolimits_{n=1}^{N}p_n\bm{k}_n,\quad \bm{c}_j=\bm{k}_j-\bar{\bm{k}}.
    \label{eq:app_key_mean_centering}
\end{equation}

Adopting Eq.~(\ref{eq:app_attention_gradient}) and Eq.~(\ref{eq:app_key_mean_centering}), we further prepare the Jacobian of $\bar{\bm{k}}$ \emph{w.r.t} $\bm{q}$ as:
\begin{equation}
\begin{split}
    \nabla_{\bm{q}}\bar{\bm{k}}
    &=\sum\nolimits_{n=1}^{N}\bm{k}_n(\nabla_{\bm{q}}p_n)^{\top} \\
    &=\frac{1}{\sqrt{d}}\sum\nolimits_{n=1}^{N}p_n\bm{k}_n (\bm{k}_n-\bar{\bm{k}})^{\top} \\
    &=\frac{1}{\sqrt{d}}(\sum\nolimits_{n=1}^{N}p_n\bm{k}_n\bm{k}_n^{\top}
    -\bar{\bm{k}}\bar{\bm{k}}^{\top}),
\end{split}
    \label{eq:app_key_mean_jacobian}
\end{equation}

Differentiating Eq.~(\ref{eq:app_attention_gradient}) to the second-order gradient and noting that $\nabla_{\bm{q}}\bm{c}_j=-\nabla_{\bm{q}}\bar{\bm{k}}$ now gives:
\begin{equation}
\begin{split}
    \nabla_{\bm{q}}^2p_j&=\frac{1}{\sqrt{d}}\nabla_{\bm{q}}(p_j\bm{c}_j)=\frac{1}{\sqrt{d}}(\bm{c}_j(\nabla_{\bm{q}}p_j)^\top + p_j\nabla_{\bm{q}}\bm{c}_j) \\
    &=\frac{1}{\sqrt{d}}(\bm{c}_j(\nabla_{\bm{q}}p_j)^\top - p_j\nabla_{\bm{q}}\bar{\bm{k}}) \\
    &= \frac{p_j}{d} (\bm{c}_j\bm{c}_j^\top - (\sum\nolimits_{n=1}^N p_n\bm{k}_n\bm{k}_n^\top - \bar{\bm{k}}\bar{\bm{k}}^\top)) \\
    &=\frac{p_j}{d}[
    (\bm{k}_j-\bar{\bm{k}})(\bm{k}_j-\bar{\bm{k}})^{\top}
    -\sum\nolimits_{n=1}^{N}p_n
    (\bm{k}_n-\bar{\bm{k}})(\bm{k}_n-\bar{\bm{k}})^{\top}
    ].
\end{split}
    \label{eq:app_attention_hessian}
\end{equation}
Restoring the window index and evaluating at $\bar{\bm{q}}_g$ gives $p_j\rightarrow p_{g,j}$ and $\bar{\bm{k}}\rightarrow\bar{\bm{k}}_g=\sum_{n=1}^N p_{g,n}\bm{k}_n$. This yields the following closed-form Hessian, which is identical to Eq.~(\ref{eq:hessian}) in Sec.~\ref{sec:tcam}.
\begin{equation}
\bm{H}_{g,j}=\frac{p_{g,j}}{d}[(\bm{k}_j - \bar{\bm{k}}_g)(\bm{k}_j - \bar{\bm{k}}_g)^\top - \sum\nolimits_{n=1}^{N} p_{g,n} (\bm{k}_n - \bar{\bm{k}}_g)(\bm{k}_n - \bar{\bm{k}}_g)^\top].
\end{equation}

\textbf{Contracting Hessian with Query Covariance.}
We first denote the diagonal entries of $\bm{\Sigma}_{g}$ by $\sigma_{g,c}^2$, where $c=1,\ldots,d$, which also corresponds to the channel-wise variances as:
\begin{equation}
    \sigma_{g,c}^2 = \frac{1}{w^2}\sum\nolimits_{i=1}^{w^2} \bm{\epsilon}_{g,i,c}^2.
\end{equation}

Since cross-channel correlations among the query deviations are typically weak, the off-diagonal covariance terms can be neglected. To further reduce sensitivity to channel-wise variance fluctuations, we approximate the covariance with an isotropic form:
\begin{equation}
    \bm{\Sigma}_g\approx \sigma_{g}^2\bm{I},\quad \sigma_{g}^2 = \frac{1}{d}\sum\nolimits_{c=1}^d \sigma_{g,c}^2.
    \label{eq:app_sigma_approx}
\end{equation}

Substituting Eq.~(\ref{eq:app_sigma_approx}) and Eq.~(\ref{eq:app_attention_hessian}) into Eq.~(\ref{eq:app_trace_approximation}), and adopting the identity $\operatorname{Tr}(\bm{x}\bm{x}^\top) = \|\bm{x}\|_2^2$, the approximated discrepancy gives:
\begin{equation}
\begin{split}
    \Delta_{g,j} &\approx \frac{\sigma_g^2}{2}\operatorname{Tr}(\bm{H}_{g,j})\\
    &=\frac{p_{g,j}\sigma_g^2}{2d}\operatorname{Tr}[
    (\bm{k}_j-\bar{\bm{k}}_g)(\bm{k}_j-\bar{\bm{k}}_g)^{\top}
    -\sum\nolimits_{n=1}^{N}p_{g,n}
    (\bm{k}_n-\bar{\bm{k}}_g)(\bm{k}_n-\bar{\bm{k}}_g)^{\top}]\\
    &=\frac{p_{g,j}}{2d} (\sigma_g^2\|\bm{k}_j-\bar{\bm{k}}_g\|_2^2-\sum\nolimits_{n=1}^{N}p_{g,n}
    \sigma_g^2\|\bm{k}_n-\bar{\bm{k}}_g\|_2^2)\\
    &\equiv \frac{p_{g,j}}{2d} (r_{g,j}-\bar{r}_g),
    \label{eq:app_discrepancy_final}
\end{split}
\end{equation}
where we define:
\begin{equation}
    r_{g,j} = \sigma_g^2\|\bm{k}_j-\bar{\bm{k}}_g\|_2^2,\quad \bar{r}_g = \sum\nolimits_{n=1}^N p_{g,n} r_{g,n}.
\end{equation}

\textbf{Calibrated Importance Score.} 
Revisiting the global calibration target in Eq.~(\ref{eq:app_calibration_target}), the approximated window-wise discrepancy in Eq.~(\ref{eq:app_discrepancy_final}) yields the calibrated window-level score:
\begin{equation}
    \tilde{p}_{g,j}=p_{g,j}+\lambda\frac{p_{g,j}}{2d}(r_{g,j} - \bar{r}_g),
    \label{eq:app_calibration_p}
\end{equation}
where $\lambda$ controls the correction strength. Finally, the overall calibrated importance score formulates:
\begin{equation}
    \tilde{s}_j = \frac{1}{G}\sum\nolimits_{g=1}^{G} \tilde{p}_{g,j}.
\end{equation}
This corresponds to Eq.~(\ref{eq:calibration}) in Sec.~\ref{sec:tcam}, augmenting the efficient pooled attention with a modest additional second-order correction to improve token discriminability for more reliable pruning.

\subsection{Derivation of the Fast Scoring Operator}
\label{sec:app_simplified_formula_derivation}

\textbf{Simplified Window-Wise Correction.} We further simplify Eq.~(\ref{eq:calibration}), equivalently given by Eq.~(\ref{eq:app_calibration_p}), in preparation for practical implementation. 

First, the centered squared distance $r_{g,j}$ can be expanded as:
\begin{equation}
    r_{g,j}
    =\sigma_g^2\|\bm{k}_j-\bar{\bm{k}}_g\|_2^2 
    =\sigma_g^2(\|\bm{k}_j\|_2^2-2\bm{k}_j^{\top}\bar{\bm{k}}_g+\|\bar{\bm{k}}_g\|_2^2).
    \label{eq:app_expand_rj}
\end{equation}
Then, given that $\sum_{n=1}^N p_{g,n}=1$ and
$\sum_{n=1}^N p_{g,n}\bm{k}_n=\bar{\bm{k}}_g$, the weighted mean distance $\bar{r}_g$ is simplified as:
\begin{equation}
\begin{split}
    \bar{r}_g
    &=\sum\nolimits_{n=1}^{N}p_{g,n}r_{g,n} \\
    &=\sigma_g^2(\sum\nolimits_{n=1}^{N}p_{g,n}\|\bm{k}_n\|_2^2
    -2(\sum\nolimits_{n=1}^{N}p_{g,n}\bm{k}_n)^{\top}\bar{\bm{k}}_g
    +\|\bar{\bm{k}}_g\|_2^2\sum\nolimits_{n=1}^{N}p_{g,n}
    ) \\
    &=\sigma_g^2(\sum\nolimits_{n=1}^{N}p_{g,n}\|\bm{k}_n\|_2^2-\|\bar{\bm{k}}_g\|_2^2).
\end{split}
    \label{eq:app_expand_rbar}
\end{equation}
Subtracting Eq.~(\ref{eq:app_expand_rbar}) from
Eq.~(\ref{eq:app_expand_rj}) and substituting the result into Eq.~(\ref{eq:app_calibration_p}) ultimately yields:
\begin{equation}
    \tilde{p}_{g,j}=p_{g,j}[1+\frac{\lambda\sigma_g^2}{2d}
    (\|\bm{k}_j\|_2^2-2\bm{k}_j^{\top}\bar{\bm{k}}_g+2\|\bar{\bm{k}}_g\|_2^2-\sum\nolimits_{n=1}^{N}p_{g,n}\|\bm{k}_n\|_2^2)].
    \label{eq:app_operator_form}
\end{equation}
This is algebraically equivalent to the TCAM formulation. It eliminates the explicit computation of $\bar{r}_g$, requiring only the window-level statistics $\bar{\bm{k}}_g$ and $u_g=\sum\nolimits_{n=1}^N p_{g,n}\|\bm{k}_n\|_2^2$, both of which can be accumulated online as in Algo.~\ref{alg:tcam_operator}.

\textbf{Exactness of Online Softmax.} To establish the basis for the tiled implementation, we first review the exactness of online Softmax. Without loss of generality, we consider a single query $\bm{q}$ within window $g$ and let $\bm{z}=\{z_j\}_{j=1}^{N}$ denote its attention logits.

Softmax is invariant to a uniform shift of its logits, \emph{i.e.} for any scalar $c$, 
\begin{equation}
    \operatorname{Softmax}(\bm{z})_j = \frac{\exp(z_j - c)}{\sum\nolimits_{n=1}^N \exp(z_n - c)}.
    \label{eq:app_softmax_invariant}
\end{equation}
In particular, choosing $c=\max_j z_j$ improves numerical stability without changing the resulting probabilities. Online Softmax exploits this property by dynamically updating the reference maximum as key blocks are processed.

Specifically, consistent with the notations in Algo.~\ref{alg:tcam_operator}, let the key sequence $\bm{K}$ be partitioned into $T_K$ blocks, each of size $B_K\times d$. After processing the first $b$ blocks, define the running maximum $m_b$ and the normalization factor $\ell_b$ as:
\begin{equation}
    m_b = \max_{j=1,\ldots,bB_K} z_j, \quad \ell_b = \sum\nolimits_{j=1}^{bB_K}\exp(z_j - m_b).
\end{equation}

When processing a new block, the running maximum changes from $m_{b-1}$ to $m_b$, and all previously accumulated terms can then be exactly rescaled to the new reference maximum using:
\begin{equation}
    \alpha = \exp (m_{b-1} - m_b),
\end{equation}
yielding:
\begin{equation}
    \ell_b = \alpha\ell_{b-1} + \sum\nolimits_{j=(b-1)B_K+1}^{bB_K} \exp (z_j - m_b).
\end{equation}
This recurrence preserves the exact normalization throughout block-wise accumulation. After the first pass till the final block, the resulting $m_{T_K}$ and $\ell_{T_K}$ recover each attention probability as:
\begin{equation}
    p_j = \frac{\exp(z_j-m_{T_K})}{\ell_{T_K}} = \operatorname{Softmax}(\bm{z})_j.
\end{equation}
Thus, tiled online computation is algebraically equivalent to standard Softmax, differing only in the accumulation procedure.

\textbf{Online Accumulation of TCAM Statistics.} Given the shift invariance of Eq.~(\ref{eq:app_softmax_invariant}), any statistic aggregated with normalized Softmax weights is likewise invariant to a uniform logit shift, \emph{e.g.} for arbitrary $\{\bm a_j\}_{j=1}^{N}$ and scalar $c$,
\begin{equation}
    \sum\nolimits_{j=1}^N p_j\bm{a}_j = \frac{\sum\nolimits_{j=1}^N\exp(z_j - c)\bm{a}_j}{\sum\nolimits_{n=1}^N \exp(z_n - c)}.
\end{equation}

Building on this, the two attention-weighted statistics required by TCAM, \emph{i.e.} $\bar{\bm{k}}_g = \sum\nolimits_{j=1}^N p_{g,j} \bm{k}_j$ and $\sum\nolimits_{n=1}^N p_{g,n} \|\bm{k}_n\|_2^2$, can be accumulated jointly within the first online pass. Concretely, after processing the first $b$ key blocks, we maintain the corresponding unnormalized running statistics:
\begin{equation}
    \bm{v}_b=\sum\nolimits_{j=1}^{bB_K}\exp(z_j-m_b)\bm{k}_j,\quad u_b=\sum\nolimits_{j=1}^{bB_K}\exp(z_j-m_b)\|\bm{k}_j\|_2^2.
\end{equation}
Both follow the same rescaling rule when the running maximum changes, \emph{i.e.},
\begin{equation}
\begin{split}
    \bm{v}_b&=\alpha_b\bm{v}_{b-1}+\sum\nolimits_{j=(b-1)B_K+1}^{bB_K}\exp(z_j-m_b)\bm{k}_j, \\
    u_b&=\alpha_b u_{b-1}+\sum\nolimits_{j=(b-1)B_K+1}^{bB_K}\exp(z_j-m_b)\|\bm{k}_j\|_2^2.
\end{split}
\end{equation}

After the final block, normalization by $\ell_{T_K}$ exactly recovers the required statistics as:
\begin{equation}
    \bar{\bm{k}}_g = \frac{\bm{v}_{T_K}}{\ell_{T_K}}=\sum\nolimits_{j=1}^{N}p_{g,j}\bm{k}_j,
    \quad
    \sum\nolimits_{n=1}^{N}p_{g,n}\|\bm{k}_n\|_2^2=\frac{u_{T_K}}{\ell_{T_K}}.
\end{equation}

With these finalized statistics, a second traversal over the key blocks recomputes the exact probabilities and directly applies the calibration, yielding the final TCAM scores. Consequently, neither the full attention matrix nor the centered-distance matrix needs to be materialized.

\end{document}